\documentclass[10pt,journal,compsoc]{IEEEtran}

\ifCLASSOPTIONcompsoc
  \usepackage[nocompress]{cite}
\else
  \usepackage{cite}
\fi

\ifCLASSINFOpdf
\else
\fi
\usepackage{times}
\usepackage{epsfig}
\usepackage{graphicx}
\usepackage{amsmath}
\usepackage{amssymb}
\usepackage{bbm}
\usepackage[breaklinks=true,bookmarks=false]{hyperref}
\usepackage{soul}
\usepackage{color}

\begin{document}
\renewcommand{\baselinestretch}{.99}
\title{Hybrid LSTM and Encoder-Decoder Architecture for Detection of Image Forgeries}

\author{Jawadul H.~Bappy,
        ~Cody~Simons,
        ~Lakshmanan~Nataraj,
	    ~B.S.~Manjunath,
        ~and~Amit K.~Roy-Chowdhury
\IEEEcompsocitemizethanks{\IEEEcompsocthanksitem J. Bappy was with the ECE department at University of California, Riverside at the time of this work. He is currently at JD.com. \protect\\ 
E-mail: {mbappy}@ece.ucr.edu
\IEEEcompsocthanksitem C. Simons, and A. Roy-Chowdhury are with the Department
of Electrical and Computer Engineering, University of California, Riverside, CA.\protect\\
E-mail: {mbappy,csimons,amitrc}@ece.ucr.edu
\IEEEcompsocthanksitem L. Nataraj is with Mayachitra Inc., Santa Barbara, CA.\protect \\ 
E-mail: nataraj@mayachitra.com
\IEEEcompsocthanksitem B.S. Manjunath is with Mayachitra Inc. and University of California, Santa Barbara, CA.\protect\\
E-mail: manj@ucsb.edu
}
}

\markboth{}%
{Shell \MakeLowercase{\textit{et al.}}: Bare Demo of IEEEtran.cls for Computer Society Journals}


\IEEEtitleabstractindextext{%
\begin{abstract} 
	With advanced image journaling tools, 
	one can easily alter the semantic meaning of an image by exploiting certain manipulation techniques such as copy-clone, object splicing, and removal, which mislead the viewers.
    In contrast, the identification of these manipulations becomes a very challenging task as manipulated regions are not visually apparent.
	This paper proposes a high-confidence  manipulation localization architecture 
	which utilizes resampling features, Long-Short Term Memory (LSTM) cells, and encoder-decoder network to segment out manipulated regions from non-manipulated ones. 
	Resampling features are used to capture artifacts like JPEG quality loss, upsampling, downsampling, rotation, and shearing.
The proposed network exploits larger receptive fields (spatial maps) and frequency domain correlation  to analyze the discriminative characteristics between manipulated and non-manipulated regions by incorporating encoder and LSTM network. Finally, decoder network learns the mapping from
low-resolution feature maps to pixel-wise predictions for image tamper localization. 

    With predicted mask provided by final layer (softmax) of the proposed architecture, end-to-end training is performed to learn the network parameters through back-propagation using ground-truth masks. Furthermore, a large image splicing dataset is introduced to guide the training process.
    The proposed method is capable of localizing image manipulations at pixel level with high precision, which is demonstrated through rigorous experimentation on three diverse datasets.
\end{abstract}

}

\maketitle

\section{Introduction}

The detection of image forgery has become very difficult
as manipulated images are often visually indistinguishable from real images. With the advent of high-tech image editing tools, an image can be manipulated in many ways. The types of image manipulation 
can broadly be classified into two categories: (1) content-preserving, and (2) content-changing  \cite{joseph2015literature}.
The first type of manipulation (e.g., compression, blur and contrast enhancement) occurs mainly due to post-processing, and they are considered as less harmful since they do not change any semantic content. 
The latter type (e.g., copy-move, splicing, and object removal) reshapes image content arbitrarily and alters the semantic meaning significantly \cite{joseph2015literature}.  The content-changing manipulations can convey false or misleading information. 
As the number of tampered images grows at an enormous rate, it becomes crucial to detect the manipulated images to prevent viewers from being presented with misleading information. 
Recently, the detection of content-changing manipulation from an image or a video has become an area of growing interest in diverse scientific and security/surveillance applications.
In this paper, we present a novel architecture to localize manipulated regions at pixel level for content-changing manipulation.
 
\begin{figure}[t]
  	\begin{center}
  		\begin{tabular}{c}
  			\includegraphics[width=.98\linewidth]{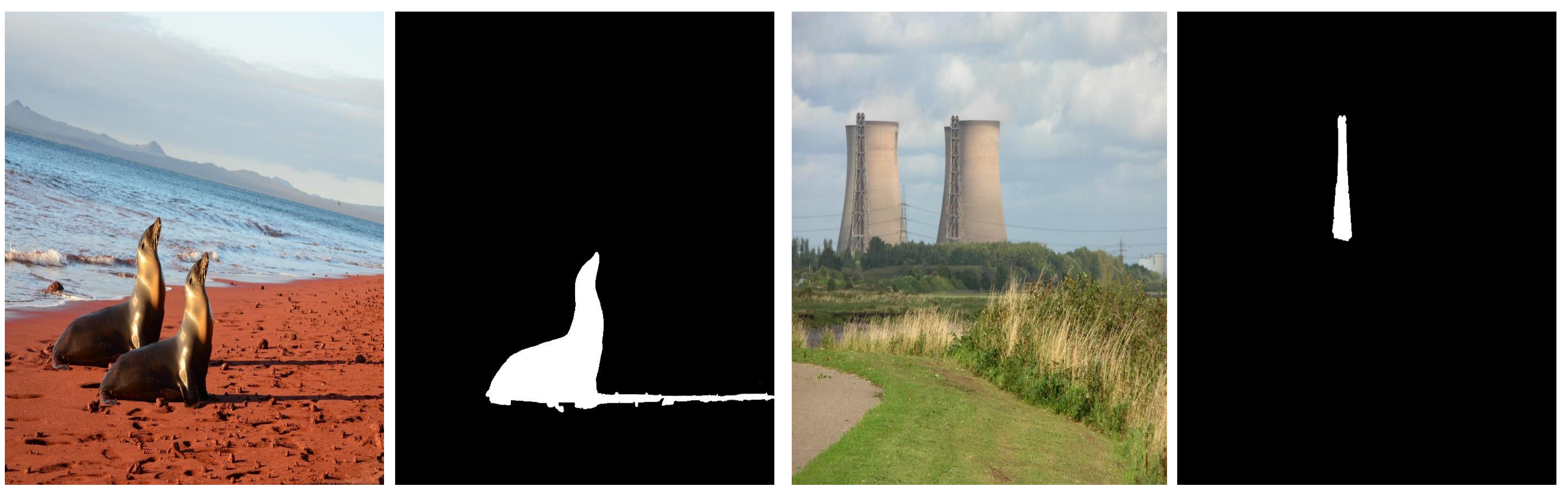}\\
  			(a) \\
  			\includegraphics[width=.98\linewidth]{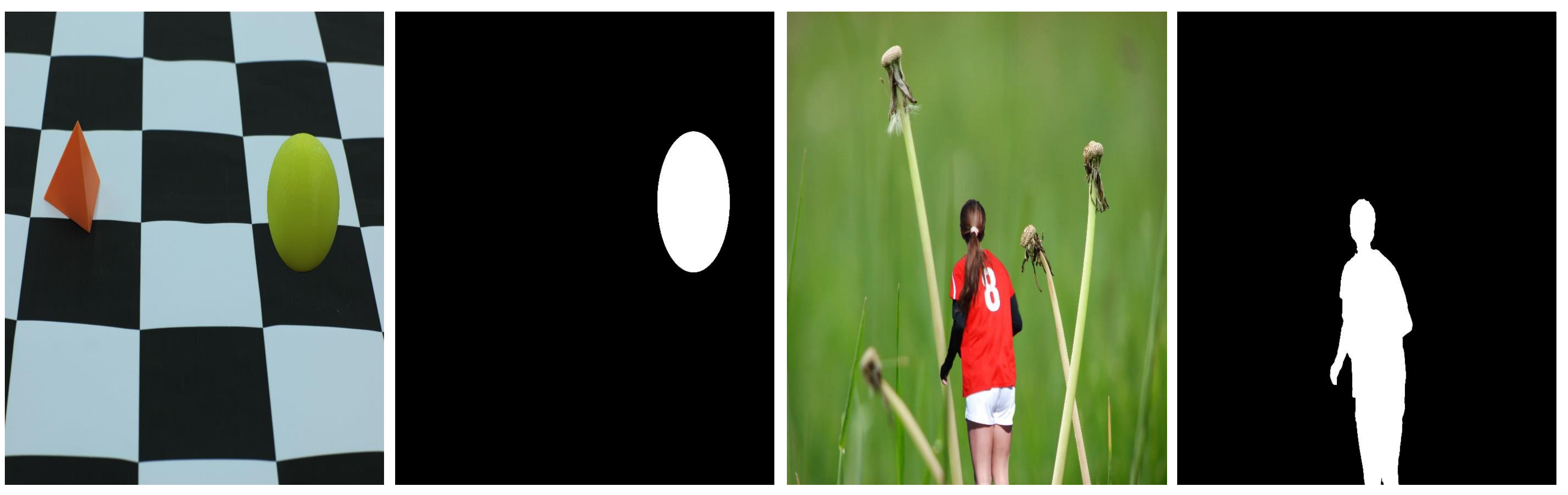}\\
  			(b) \\
  			\includegraphics[width=.98\linewidth]{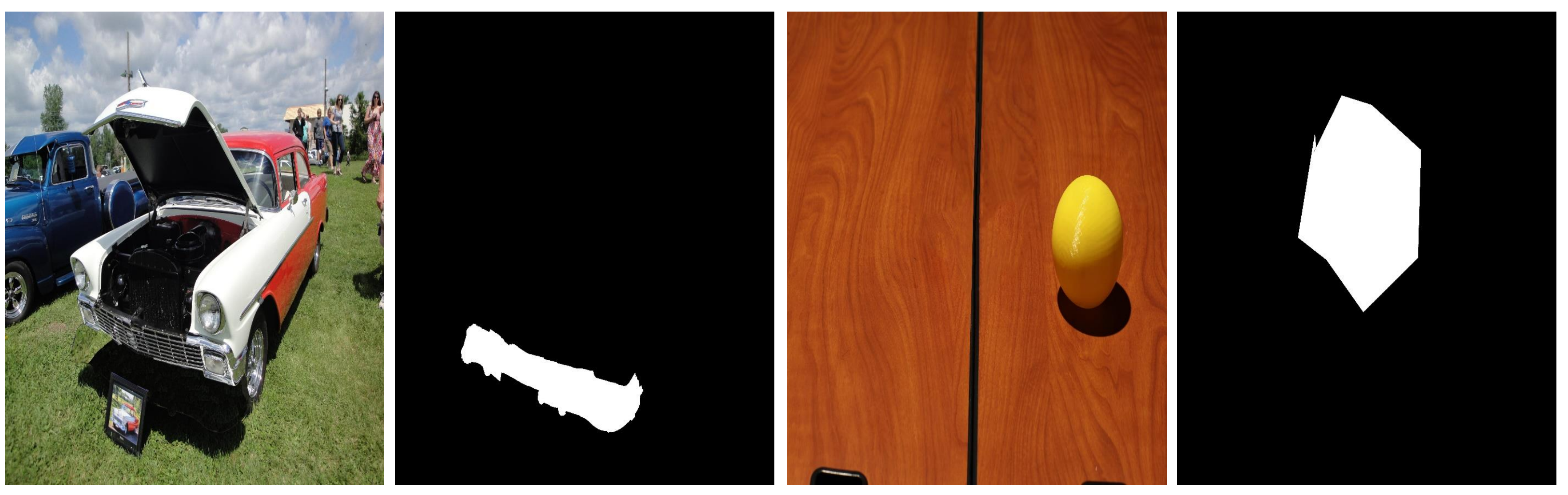}\\
  			(c)
  		\end{tabular}
  	\end{center}
  	\vspace{-2mm}
   	 \caption{The figure demonstrates some examples of content-changing manipulations.
   	   (a), (b), (c) illustrate copy-clone, splicing and object removal techniques to manipulate an image. First and third columns are tampered images and corresponding ground-truth masks are shown in second and fourth columns.
}
  	\label{motivation}  
  	\vspace{-2mm}
  \end{figure}

Over the past decades, there have been many works to classify image manipulation, i.e., whether an image is tampered or not \cite{muhammad2012passive, bianchi2012image, li2017image, jaberi2014accurate, pun2015image,ryu2013rotation, ferrara2012image}.  However, only few works \cite{bappy2017exploiting, bondi2017tampering}  attempt to localize image manipulation at pixel level. Some recent works \cite{bunk2017detection,fan2015general, liu2017image} address the localization problem by classifying patches as manipulated.
The localization of image tampering is a very challenging task as well-manipulated images do not leave any visual clues, as shown by the following examples in Fig.~\ref{motivation}. 
In Fig.~\ref{motivation}(a), 
copy-move manipulation is illustrated where one object is copied to another region of the same image leading to two similar objects, one originally present, and another manipulated. However, only the latter needs to be identified. Fig.~\ref{motivation}(b) illustrates object splicing manipulation, where an object from a donor image has been spliced into other image. As another example, if an object is removed as shown in Fig.~\ref{motivation}(c), the region may visually blend into the background, but needs to be identified as manipulated.


Most of the state-of-the-art image tamper classification approaches utilize the  frequency domain characteristics and/or statistical properties of an image \cite{li2007sorted,luo2006robust,mahdian2007detection,wang2014exploring}. 
The analysis of artifacts by multiple JPEG compressions is also utilized in \cite{chang2013forgery, wang2014exploring}  to detect
manipulated images, which are applicable only to the JPEG formats.
In \cite{nataraj2010improving,nataraj2009adding}, noise has been added to the JPEG compressed image in order to improve the performance of resampling detection.
In computer vision, deep learning has shown promising performance in different visual recognition tasks such as object detection \cite{girshick2015fast}, scene classification \cite{zhou2014learning},  and semantic segmentation \cite{long2015fully}. Some recent deep learning based methods such as stacked auto-encoders (SAE) \cite{zhang2016image} and convolutional neural networks (CNN) \cite{rao2016deep,bayar2016deep,chen2015median} have also been applied to detect/classify image manipulations. 
In media forensics,
most of the existing forgery detection approaches focus on identifying  a specific tampering method, such as copy-move \cite{cao2012robust,hashmi2014copy,li2015segmentation},  and splicing \cite{manu2015visual}.  Thus, one approach might not do well on other  types of tampering. 
Moreover, it seems unrealistic to assume that the type of manipulation will be known beforehand. Our recent paper \cite{bappy2017exploiting}, upon which this particular work builds, presents a general detection architecture for different content-changing manipulations.

 Unlike semantic object segmentation where all meaningful regions (objects) are segmented, the localization of image manipulation focuses only the possible tampered region which makes the problem even more challenging.
 In computer vision, recent advances in semantic segmentation methods \cite{zheng2015conditional, long2015fully, badrinarayanan2017segnet} are based on convolutional neural networks (CNN).
 In \cite{zheng2015conditional}, a fully convolutional network is 
 utilized to analyze the content of the objects and shape of a region by extracting the hierarchical features at different levels.
 In object detection \cite{girshick2015fast} and segmentation \cite{long2015fully, badrinarayanan2017segnet}, CNN based architectures demonstrate very promising performance in understanding visual concepts by analyzing the content of different regions. In contrast to semantic segmentation, manipulated regions could be removed objects, or copied objects from other parts of the image. Well-manipulated images usually   show strong resemblance between fake and  genuine objects/regions (i.e. content is similar) \cite{rao2016deep}. Even though CNN generates spatial maps for different regions of an image,
 it can not generalize some other artifacts created by different manipulation techniques. Thus, the localization of manipulated regions with only CNN based architecture may not be the best strategy. In our earlier work \cite{bappy2017exploiting}, we compared with some recent  semantic segmentation approaches \cite{zheng2015conditional, long2015fully} that do not perform well for copy-clone and object removal type of manipulations.

Image tampering creates some artifacts, e.g., resampling, compression, shearing, which are better captured by resampling features \cite{ryu2014estimation, bunk2017detection, feng2012normalized}. 
In \cite{bunk2017detection}, a Long short-term memory
(LSTM) based network is presented in order to classify manipulated patches where resampling features are utilized as an important signature.
The authors trained six classifiers to detect six different types of resampling (e.g.,  JPEG quality thresholded above
or below 85, upsampling, downsampling, rotation clockwise,
rotation counterclockwise, and shearing.
Resampling introduces periodic correlations among pixels due to interpolation. As convolutional neural networks exhibit robust  translational invariance to generate spatial maps for the different regions of the image, and certain artifacts are well-captured in resampling features, both can be exploited in order to localize manipulated regions.

Towards the goal of localizing manipulated regions in an image, \emph{we present a unified architecture that exploits resampling features, LSTM network, and encoder-decoder architectures in order to learn the pixel level localization of manipulated image regions}.
Given an image, we divide into several blocks/patches and then resampling features (as discussed in Sec.~\ref{resampling}) are extracted from each block. 
LSTM network is utilized to learn the correlation between manipulated and non-manipulated blocks at frequency domain. 
We utilize and modify encoder-decoder network as presented in \cite{badrinarayanan2017segnet}   to capture spatial information.
Each encoder generates feature maps with varying size and number. The feature maps from LSTM network and the encoded feature maps from encoders are embedded before going through the decoder.   
We perform end-to-end training to learn the parameters of the network through back-propagation using ground-truth mask information. 
As deep networks are data hungry, a large number of images are synthesized to augment the training data..
The proposed model shows promising results in localizing manipulated regions at the pixel level,  which is demonstrated on different challenging datasets.

 \begin{figure*}[t]
 	\begin{center}
 		\begin{tabular}{c}
 			\includegraphics[width=.8\linewidth]{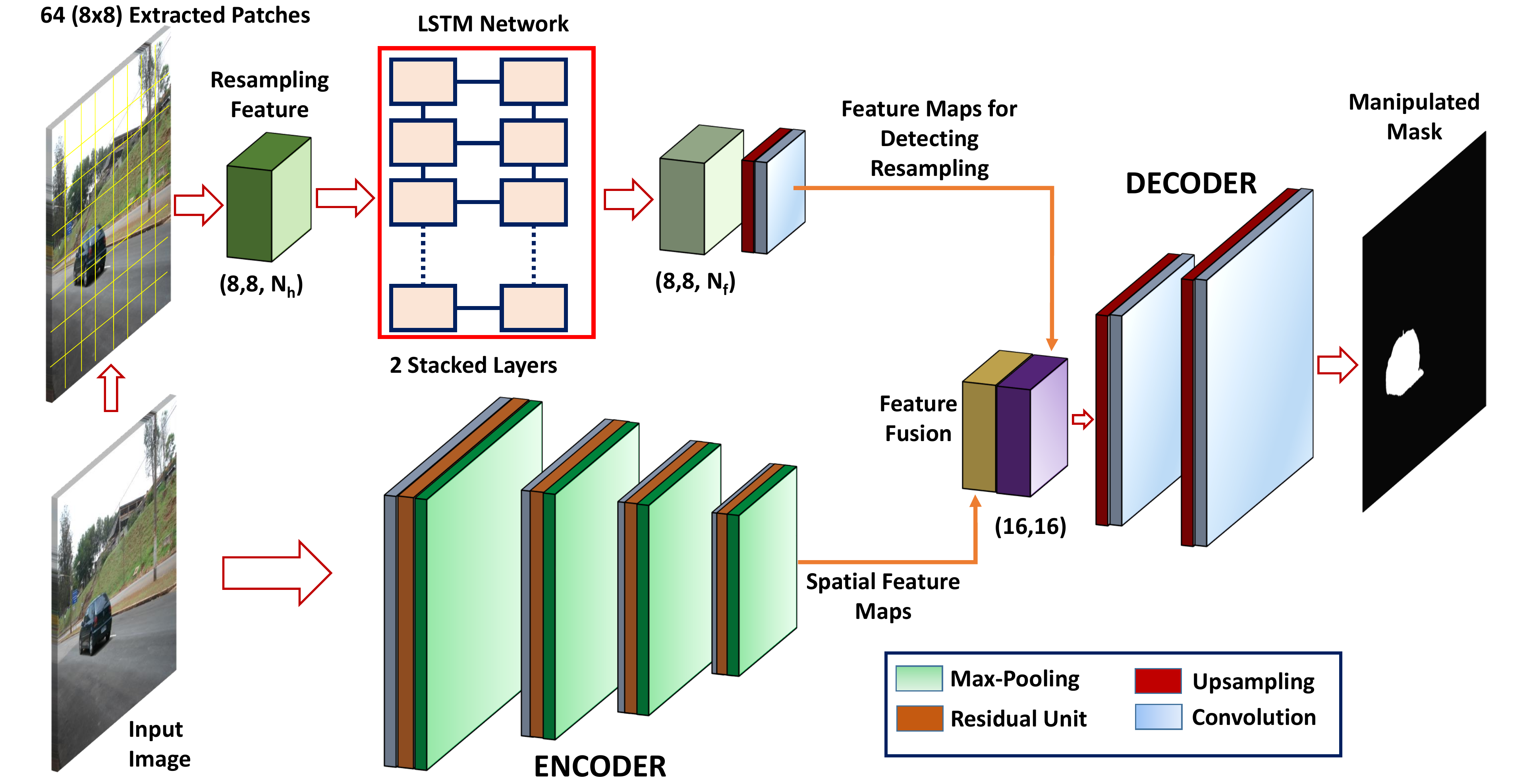}\\
 		\end{tabular}
 	\end{center}
 	\vspace{-2mm}
 	\caption{
 		Overview of proposed framework for localization of manipulated image regions. 
 			}
 	\label{ovFrame}  
 	\vspace{-2mm}
 \end{figure*}

\subsection{Approach Overview}
Given an image, our goal is to localize the manipulated regions at a pixel level. The proposed framework is shown in Fig.~\ref{ovFrame}.
Our network can be divided into three parts- (1) LSTM network with resampling features, and (2) convolutional encoder, and (3) decoder network. 

For the first part, we divide image into patches. For each patch, resampling features \cite{bunk2017detection} have been extracted. With extracted resampling features, we use Hilbert curve (discussed in Sec.~\ref{hilbert}) to determine the ordering of the patches to feed into LSTM cells. 
We allow LSTM cells to learn the transition between manipulated and non-manipulated blocks in the frequency domain. Finally, feature maps are generated from the LSTM cell output, which will be combined with the feature maps from the encoder.
An encoder consists of residual block, batch normalization and  activation function. At each residual block, two convolutions are performed with shortcut connection. After each residual unit, max-pooling operation is performed which gives translation invariance.

Our next step is to design a decoder that can provide finer representation of different regions in a mask. 
We combine both spatial features from encoder and output features from LSTM to understand the nature of manipulation. Then, these features are taken as input to the decoder.
Each decoder follows basic operations like upsampling, convolution, batch normalization and activating feature maps (using activation function). The decoders help learn the finer details of the manipulated and non-manipulated classes.
Finally, 
a softmax layer is used to predict manipulated pixels against non-manipulated ones.
With the ground-truth mask of manipulated regions we perform end-to-end training to classify each pixel.  We compute cross entropy loss, which is then minimized by utilizing back-propagation algorithm.  
After optimization, we find the optimal set of parameters for the network, that will be used to predict manipulated regions given a test set.

 \subsection{Main Contributions} 
 Our \textit{main contributions} are as follows.
 
 $\bullet$ We propose a novel localization framework that exploits both frequency domain features and spatial context in order to localize manipulated image regions, which makes our work significantly different than other state-of-the-art methods.

$\bullet$ Unlike most of the existing works where patches are used as input, we consider image as input so that we can utilize global context. Our architecture is able to localize manipulated region with high confidence as demonstrated on three datasets.

$\bullet$ We present a new dataset for image tamper localization task  that includes a large number of images with ground-truth binary mask.
This dataset is larger than current publicly available datasets such as IEEE Forensics \cite{2013-ifc-challenge} and COVERAGE \cite{wen2016}. It will also help train deeper networks for image tamper classification or localization tasks.  

This work builds upon our earlier paper \cite{bappy2017exploiting}, but with significant differences.
 First, the method presented in the paper \cite{bappy2017exploiting} exploits low level features such as tampered edges, as evidence of tampering, which cannot always detect the entire tampered regions. The proposed method exploits an encoder and an LSTM network to extract spatial feature maps and frequency domain features respectively in order to localize manipulated regions. In the proposed method, the encoder provides larger receptive fields by exploiting multiple convolutional layers which allow the network to identify large manipulated regions.
Second,
we consider the image as the input, instead of patches, which helps the network to learn more meaningful  context, i.e., intra-patch and inter-patch correlation. Third, unlike \cite{bappy2017exploiting}, we utilize resampling features in our network that captures the characteristics of  different artifacts due to image transformation such as JPEG quality above or below a threshold, upsampling, downsampling, rotation clockwise,  rotation  counterclockwise,  and  shearing. Fourth, we present a large image splicing dataset which can be used to train a deep neural network for the task of manipulation. We show  the comparison against existing dataset in experimental analysis. 

\section{Related Work}
 In media forensics, there have been lot of efforts to detect different types of manipulations such as resampling, JPEG artifacts, and content-changing manipulations.
In this section, we will briefly discuss some of the existing works for detecting image forgeries.

In last few years, several methods have been proposed to detect resampling in digital images \cite{ryu2014estimation, Nataraj10-345,  babak-radon, popescu2005exposing, feng2012normalized}.
Most of the  approaches exploit linear or cubic interpolation.
In \cite{ryu2014estimation},   periodic properties of interpolation by the second-derivative of the transformed image have been utilized for detecting image manipulation.
In \cite{Nataraj10-345},  an approach was presented to identify resampling on JPEG compressed images where noise was added before passing the image through the resampling detector; it was shown that adding noise aids in detecting resampling. In \cite{feng2011energy,feng2012normalized}, a feature was generated from the normalized energy density and then SVM was used to robustly detect resampled images. 
Some recent approaches \cite{golestaneh2014algorithm, kwon2015efficient} have been proposed to reduce  JPEG artifacts produced by compression techniques. In \cite{amerini2011sift,verdoliva2014feature}, feature based methods have been presented in order to detect manipulation in an image.
Many methods have been proposed to detect seam carving \cite{sarkar2009detection,fillion2010detecting,liu2015improved} and inpainting based object removal \cite{wu2008detection,chang2013forgery,liang2015efficient}. 
Several approaches exploit JPEG blocking artifacts to detect tampered regions~\cite{lin2009fast,farid2009exposing,luo2010jpeg,bianchi2011improved,bianchi2012image}. 
Some recent works  \cite{li2015segmentation,kakar2012exposing, jaberi2014accurate,al2013passive} focus on identifying and localizing copy-move manipulation.
In \cite{li2015segmentation}, the authors used an interesting segmentation based approach to detect copy-move forgeries. 
They first divided an image into semantically independent patches and then performed keypoint matching among these patches.
In \cite{cozzolino2015efficient}, a patch match algorithm was used to efficiently compute an approximate nearest neighbor field over an image.
They further used invariant features such as Circular Harmonic transforms and showed robustness over duplicated blocks that have undergone geometrical transformations.
In \cite{manu2015visual}, an image splicing technique was presented using visual artifacts. 
In \cite{muhammad2014image}, the steerable pyramid transform (SPT) and the local binary pattern (LBP) were utilized to detect image forgeries. The paper \cite{guillemot2014image} highlights the recent advances in image manipulation and also discusses the process of restoring missing or damaged areas in an image. In \cite{ansari2014pixel}, a review on different image forgery detection techniques is presented. 

Recently, there has been a growing interest to detect image manipulations by applying different computer vision and machine learning algorithms \cite{nguyen2018modular, rahmouni2017distinguishing}.  
In semantic segmentation,  many deep learning  architectures  \cite{long2015fully,zheng2015conditional,badrinarayanan2017segnet} have been proposed, which surpass previous state-of-the-art approaches by a large margin in terms of accuracy. Most of the deep networks \cite{long2015fully,badrinarayanan2017segnet} are based on  Convolutional Neural Networks (CNNs), where hierarchical features are exploited at different layers in order to learn the spatial map for semantic region. In \cite{long2015fully},  a classification-purposed CNN is transformed into fully convolutional one by replacing fully connected layers to produce spatial heatmaps. Finally, a deconvolution layer is used to upsample the heatmaps to generate dense per-pixel labeling. 
SegNet \cite{ badrinarayanan2017segnet} designs a decoder to efficiently learn the low-resolution heatmaps for pixel-wise predictions for segmentation.
In \cite{kendall2015bayesian, chen2018deeplab}, the fully connected pairwise CRF is  utilized as a post-processing step to refine the segmentation result. In
\cite{pinheiro2016learning}, skip connection is exploited to perform
late fusion of feature maps for making independent predictions
for each layer and merging the results. 
In ReSeg \cite{visin2016reseg}, Gated Recurrent
Units (GRUs) and upsampling have been used to obtain the segmentation mask.

Recent efforts, including \cite{bayar2016deep, bayar2017design,rao2016deep,bunk2017detection,mohammed2018boosting} in the manipulation detection task, exploit deep learning based models.
These tasks include detection of generic manipulations \cite{bayar2016deep, bayar2017design}, resampling  \cite{bayar2017resampling}, splicing \cite{rao2016deep}, and bootleg \cite{buccoli2014unsupervised}.
In \cite{qian2015deep}, the authors propose Gaussian-Neuron CNN (GNCNN) for steganalysis.
A deep learning approach to identify facial retouching was proposed in \cite{bharati2016detecting}.
In \cite{zhang2016image}, image region forgery detection has been performed using stacked auto-encoder model. 
In \cite{bayar2016deep},  a new form of convolutional layer is proposed to learn the manipulated features  from an image. 
In computer vision, deep learning has led to significant performance gain in different visual recognition tasks such as image classification \cite{zhou2014learning}, and semantic segmentation \cite{long2015fully}.  
The deep networks extract hierarchical features to represent the visual concept, which is useful in object segmentation. Most of the architectures are based on Convolutional Neural Network (CNN), which provides spatial maps relevant to manipulated regions. However, we can also exploit resampling features that distinguish other artifacts.  Since both spatial context and resampling are important attributes to localize manipulated regions from image, we present an unique network that exploits both of the features.

\section{Network Architecture Overview}
\label{LSTM-Conv}

\begin{figure*}[t]
		\begin{center}
	\begin{tabular}{c}
		\includegraphics[height=.4\linewidth,width=.94\linewidth]{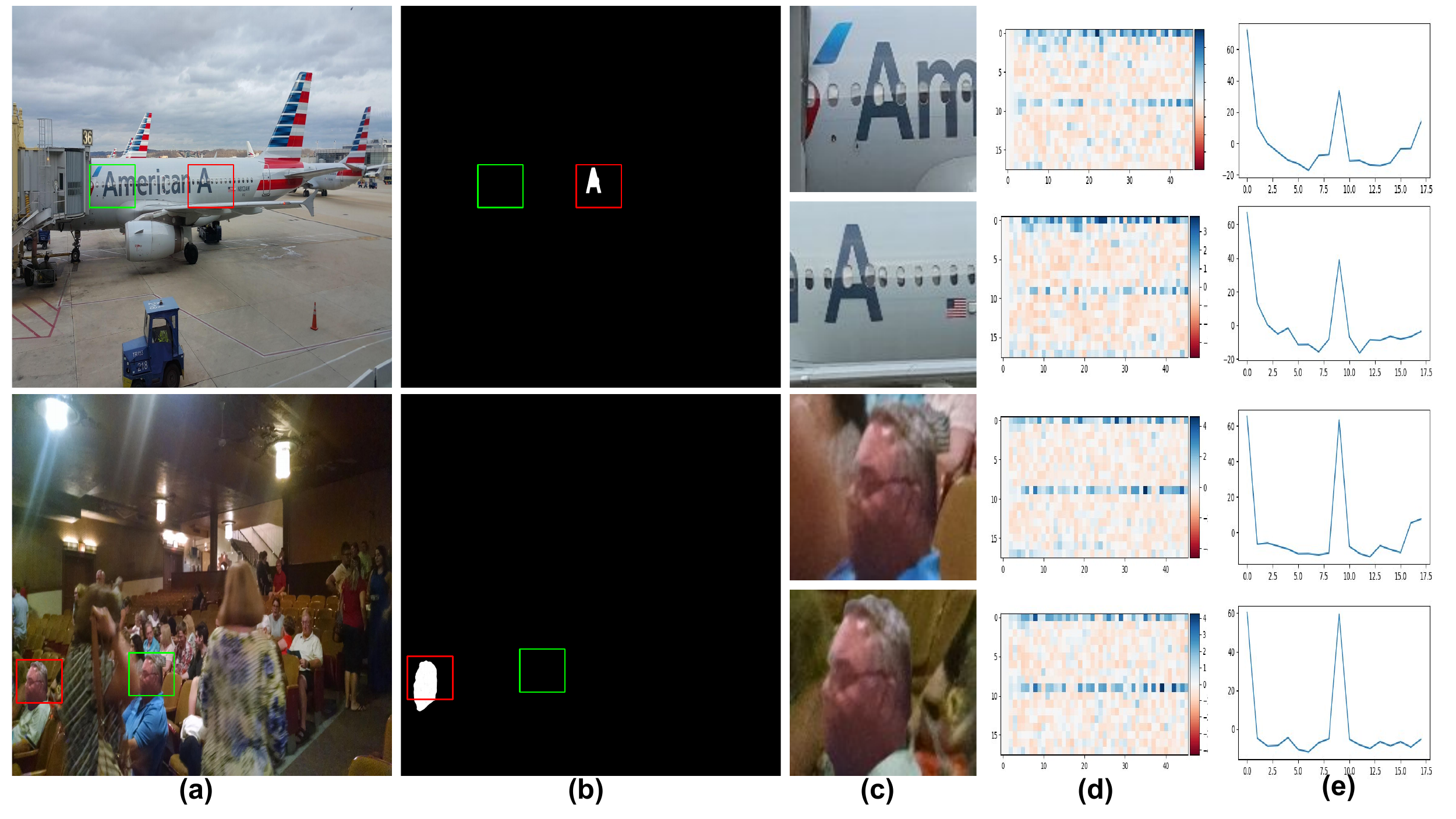} \\
	\end{tabular}
	\end{center}
	\vspace{-2mm}
	\caption{
(a) Examples of manipulated images from the NIST dataset \cite{2016-nimble-dataset}. (b) Corresponding ground-truth
masks for the manipulated images in column (a), green for non-manipulated patches and red for
manipulated ones. (c) The patches extracted from the corresponding image, with the top containing no manipulation and the bottom one
containing some manipulations. (d) Radon transform of two patches from the manipulated and non-manipulated images.
(e) Sum along the columns of the radon transform. Here, we can see that the non-manipulated patch exhibits high magnitude at the right
of the curve, whereas low value is observed in the same region for the manipulated patch. Differences such as these are seen across other manipulated and non-manipulated patches.
	}
	\label{boundary}  
	\vspace{-3mm}
\end{figure*}
Our main goal of this work is to localize image manipulations at pixel level. 
Fig.~\ref{ovFrame} shows our overall framework. The whole network can be divided into three parts - (1) LSTM network with resampling features, and (2) Encoder, and (3) Decoder network. 
Convolutional neural network (CNN) architectures extract meaningful spatial features for object segmentation, which could also be useful to localize manipulated objects.  Even though spatial feature maps are crucial to classify each pixel,  solely using CNNs in the image domain does not usually perform well in identifying image manipulations. It is simply because there are certain manipulations like upsampling, downsampling, compression,  which are well-captured in the frequency domain. Thus, we use resampling features from the extracted patches of an image.  These resampling features are considered as input to the LSTM network which learns the correlation between different patches. 
An encoder architecture is also utilized to understand the spatial location of manipulated region. 
Before decoder network, we utilize the meaningful features by exploiting both spatial and frequency domain. 
Finally, we use decoder network to obtain finer representation of binary mask to localize tampered region from low-resolution feature maps. 
In order to develop encoder-decoder network, we utilize convolutional layers, batch normalization, max-pooling and upsampling. 
Next, we will discuss the technical details of our proposed architecture for image tamper localization. 

\subsection{LSTM Network with Resampling Features}
\subsubsection{Resampling Features}
\label{resampling}
The typical content-changing manipulations are copy-clone, splicing and object removal, which are difficult to detect. In general, these manipulations distort the natural statistics at the boundary of tampered regions.
In~\cite{babak-radon}, the method of resampling detection using Radon transform is presented. 
Laplacian filter along with Radon transform is exploited  in order to extract resampling features given a patch. 
We will also follow a similar procedure for extracting resampling features. 
To illustrate how Radon transform captures resampling characteristics, we provide  two examples to highlight the difference in statistics between manipulated and non-manipulated patches as shown in Fig.~\ref{boundary}~(c), with the top row containing no manipulation and the bottom row containing some manipulations  due to resampling. 
Fig.~\ref{boundary}~(d,e) illustrates the radon transform of two patches from  the manipulated and non-manipulated regions and their sum along the columns.
Though the differences are subtle, we can see that there is a pattern for manipulated patches which is different from those of non-manipulated patches.
Given an image, we first extract $64$ ($8 \times 8$) non-overlapping patches. As input image has size of 256x256x3, the dimension of each patch would be 32x32x3. Then, the square root of  magnitude of $3 \times 3$ Laplacian filter is used to produce  the magnitude of linear predictive error for each extracted patch as presented in \cite{bunk2017detection}. 
As resampling signal has periodic correlations in the linear predictor error, we apply the Radon transform to accumulate errors along various angles of projection. In our experiment, we use $10$ angles. Finally, we apply 
Fast Fourier Transform (FFT) to find the periodic nature of the signal. 
In general, these resampling features are capable of capturing
different resampling characteristics- JPEG quality thresholded above
or below a threshold, upsampling, downsampling, rotation clockwise,
rotation counterclockwise, and shearing (in an affine
transformation matrix).
In order to reduce computational burden, we resize images to $256 \times 256 $ which might introduce some additional artifacts such as degradation in image quality factor, shearing, upsampling, downsampling. 
In \cite{bunk2017detection}, resampling features are used to classify 
these artifacts.
In this work, we also utilize resampling features, which gives us robust performance. 
Unlike \cite{bunk2017detection}, where resampling features are considered for patch classification, we perform localization at pixel level. 
There is a tradeoff in selecting the patch size: resampling
is more detectable in larger patch sizes because the
resampling signal has more repetitions, but small manipulated
regions will not be localized that well.
In \cite{bunk2017detection}, resampling features are extracted from $8 \times 8$ block. On the other hand, we choose $32 \times 32$ small patches from an image to extract resampling features that capture more information.
The major motivation of utilizing the resampling features for patches is to characterize the local artifacts due to different types of manipulations. 

  \begin{figure}[t]
  	\begin{center}
  		\begin{tabular}{c}
  			\includegraphics[width=.98\linewidth]{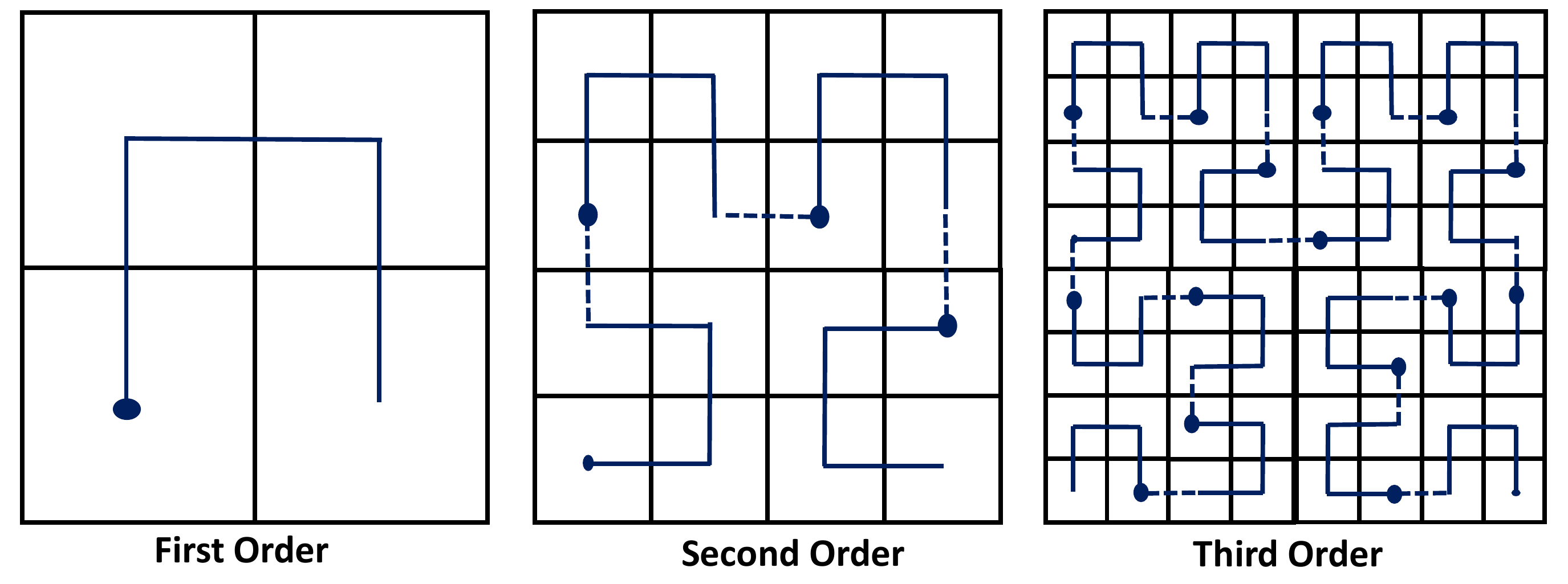}\\
  			
  		\end{tabular}
  	\end{center}
  	\vspace{-2mm}
   	 \caption{The figure illustrates Hilbert curves for different orders. In this work, third order curve has been exploited.
}
  	\label{hilb}  
  	\vspace{-2mm}
  \end{figure}
  
\subsubsection{Hilbert Curve}
\label{hilbert}
Long-Short Term Memory (LSTM) is commonly used in tasks where sequential information exists. The performance of LSTM highly depends on the ordering of the patches (sequence of the extracted patches). One can consider horizontal or vertical directions, but these orderings do not capture local information well. For example, if we were to iterate horizontally over the rows of patches, then patches that neighbor each other vertically will be separated by an entire row of patches. Due to this long time lag, LSTM can not correlate well between these patches. If we were to iterate vertically over the columns we would face similar issues. 

In order to better preserve the spacial locality of the patches, we use space-filling curve which is commonly used to reduce multi-dimensional problem to a one-dimensional \cite{breinholt1998algorithm}. 
The Hilbert curve  has been shown to outperform many other curves in maintaining the spatial locality, when transforming from a multi-dimensional space to a one-dimensional space \cite{moon2001hilbert}.
The major advantage of Hilbert curve is in applications where 
the coherence between neighboring patches/blocks is important \cite{voorhies1991space}.
Fig.~\ref{hilb} shows the process of how Hilbert curve works.
The basic elements of the Hilbert curves can be divided into `cups' (a square with one open side) and `joins' (a vector that joins two cups) \cite{voorhies1991space}. Every cup has two end-points - (a) entry point, and (b) exit point. From Fig.~\ref{hilb}(left), we can see that a single cup represents a first order Hilbert curve which fills a $2 \times 2$ space.
The second order Hilbert curve contains four cups, which are linked together by three joins as shown in  Fig.~\ref{hilb}(middle). 
A third order Hilbert curve  repeats the process by dividing into four parts, each of these parts  contains second order Hilbert curve. Finally, the four parts are connected by three joins.
 So, the main mechanism is to divide a plane into four parts, each of these parts into four parts, and so on.  
As we have total $64$ ($8 \times 8$) blocks  extracted from an image, we require three recursive dividing of the plane. After ordering the patches with Hilbert curve, LSTM network is utilized. We empirically observe that this ordering technique helps improve the performance of localization.

\subsubsection{Long-Short Term Memory (LSTM) Network}
\label{LSTM}
LSTM network is well-known for processing sequential data in different applications such as language modeling, machine translation, image captioning, and hand writing generation. 
In computer vision,  LSTM network has been successfully used  to capture the  dependency among a series of pixels \cite{pinheiro2014recurrent,byeon2015scene}. The key insight of using LSTM for detecting image manipulations is \textit{to learn the boundary transformation between different blocks, which provides discriminative features between manipulated and non-manipulated regions}.

In \cite{bappy2017exploiting, bunk2017detection}, LSTM network is utilized in order to learn the transition (change) between manipulated vs non-manipulated blocks by feeding the blocks into an LSTM network.
In \cite{bunk2017detection}, the authors propose a patch classification framework where frequency domain features are extracted from $8 \times 8$ block before LSTM network.
The method could be more effective by considering larger block size.
Unlike these approaches, we divide an image into several patches, and extract rasampling features as discussed in Sec.~\ref{resampling} from $32 \times 32$ size of patch that are taken as input to the LSTM network.

After extracting resampling features for each patch, we use Hilbert curve (discussed in Sec.~\ref{hilbert}) to determine the ordering of the patches. Then, we feed the resampling features extracted from patches into LSTM cells in a sequential manner.  LSTM network computes
the logarithmic distance of patch dependency by feeding each patch to each cell. 
The LSTM cells learn the correlation among neighboring patches. 
In this paper, we use $ 2 $ stacked layers, and  $ 64 $ time steps in LSTM network. We obtain $64$ dimensional feature vector from each time step in the last layer. Then, we project the vector generated by  LSTM network to $N_f$ features maps. Let us denote a feature vector $F_l \in \mathcal{R}^{ 1 \times N_h}$ produced by $l^{th}$ time step of LSTM network. We represent the projected vector as $F_l^\prime$ with $N_f$ dimension. In order to obtain the output $O_l$, we introduce  a weight matrix $W_l$ ($\in \mathcal{R}^{N_h \times N_f}$) which transforms from $F_l$ to $F_l^\prime$. The vector  $F_l^\prime$ can be written as 

\begin{equation}
    F_l^\prime=F_l.W_l+B_l.
\end{equation}
Here, $B_l$ is bias with $N_f$ dimension.
Each time step of LSTM  network actually provides the transformed feature for  each of the extracted patches from input image. Finally, we obtain $64 \times N_f$ size matrix for $64$ patches. In our experiment, we choose $N_h=128$ and $N_f=64$. 
Next, we carefully choose the ordering of the cell outputs in order to preserve the spatial information. Then, we reshape  the  $64 \times N_f$ matrix 
to $8\times 8\times N_f$, where first two dimensions represent the location of the patch as shown in Fig.~\ref{ovFrame}.

{\bf LSTM Cell Overview.}
Information flow between the LSTM cells is controlled by three gates: 
 (1)  input gate, (2) forget gate, and (3)  output gate. Each gate has a value ranging from zero to one, activated by a sigmoid function.
 Let us denote cell state and output state as $C_t$ and  $z_t$ for current cell $t$. 
Each cell produces new
candidate cell state $\bar{\mathcal{C}}_t $. Using the previous cell state $ \mathcal{C}_{t-1} $ and $\bar{C}_t$, we can write the updated cell state $ \mathcal{C}_t $  as  
\begin{equation}
\label{cell_state}
\mathcal{C}_t=f_t \circ \mathcal{C}_{t-1}+i_t \circ \bar{\mathcal{C}}_t
\end{equation}
Here, $ \circ $ denotes the $pointwise$ multiplication. 
Finally, we obtain the output of the current cell $ h_t $, which can be represented as 
\begin{equation}
\label{out_state}
z_t = o_t\circ tanh(\mathcal{C}_t)
\end{equation}
In Eqns.~\ref{cell_state} and ~\ref{out_state}, $i,f,o $ represent input, forget and output gates.

\subsection{Encoder Network}
Our main objective is to design an efficient architecture
for pixel-wise tamper region segmentation. We use convolutional layers to design the encoder which allows the network to understand appearance, shape and  the spatial-relationship (context) between
manipulated and non-manipulated classes. 
In \cite{badrinarayanan2017segnet, long2015fully, chen2018deeplab}, some  deep architectures are presented where convolutional layers are utilized in order to produce spatial heatmaps for semantic segmentation. As spatial information is very important to localize manipulated regions, we also incorporate convolutional layers into our framework. We exploit and modify encoder-decoder architecture as presented in \cite{badrinarayanan2017segnet}. The encoder component is  similar to CNN architecture except the fully connected layers. 

Convolutional Network (ConvNet) consists of different layers, where each 
layer of data  is a three-dimensional
array of size $h \times w \times c$, where $h$ and $w$ are height and width of the data,
and $c$ is the dimension of the channels. Each layer of convolution involves learnable filters with varying size.  The filters in convolutional layer will create feature maps that are connected to the local region of the previous layer. In the first layer, image is taken as input with dimension of $ 256 \times 256 \times 3 $ (width, height, color channels).

The basic building block of each encoder utilizes convolution, pooling, and activation functions. We use residual unit \cite{he2016deep} for each encoder. Residual block takes advantage of shortcut connections that are parameter free.
The main advantage of using residual unit is that it can easily optimize the residual mapping and more layers are trainable. Let us consider an input to the residual unit is $y$, and the mapping from input to output of the unit is $\mathcal{T}(.)$. The output of residual unit would be $\mathcal{T}(y)+y$ in the forward pass.  In each convolutional layer, we use kernel size of $ 3\times 3 \times d$, where $ d $ is the depth of a filter. We use different depth for different layers in the network. In encoder network, the number of filters are generally in increasing order.  
In this work, we utilize $32, 64, 128,$ and $256$ feature maps in first, second, third and fourth layer of encoder architecture respectively.

Each residual unit  in the encoder produces a set of feature maps. We utilize batch normalization \cite{ioffe2015batch} at each convolutional layer. Batch normalization is robust to covariance shift. 
As an activation function, we choose rectified linear unit (ReLU) \cite{nair2010rectified} that can be represented as $max(0,x)$. At the end of each residual unit, max-pooling  with stride $2$ is performed, which reduces the size of feature maps by a factor of $2$. Unlike \cite{bappy2017exploiting}, we exploit max-pooling \cite{krizhevsky2012imagenet} at each layer as it provides  translation invariance. Each max-pooling operation introduces a loss of spatial resolution (i.e., boundary details) of the feature maps. The loss in boundary detail can be compensated by using decoder which is introduced in \cite{badrinarayanan2017segnet}, and discussed next.

\subsection{Decoder Network}

In \cite{long2015fully}, a decode technique is proposed that requires encoder feature maps to be stored during prediction. This process might not be applicable in real-life as it requires intensive memory.
In this paper, we follow  a decoding technique that is  presented in \cite{badrinarayanan2017segnet}. 
In \cite{badrinarayanan2017segnet}, the advantage of using decoder has been discussed in details.
The key part is the decoder which replaces the fully connected layers. The decoder decodes the feature output from encoder. 
As encoder-decoder is primarily developed for semantic object segmentation \cite{badrinarayanan2017segnet}, we exploit and tune this network in order to segment manipulated objects. 
In the upsampling step, no learnable parameters are involved. Different multi-channel filters are utilized which are convolved with the upsampling heatmaps (coarse representation) to create dense maps.
Each decoder follows basic operations - upsample, convolution, and batch normalization. Each decoder first performs upsampling of the feature maps learned at previous layer. Following that, convolutional operation and batch normalization are performed. We employ $3 \times 3$ size kernel for  decoder network. 
In our decoder,  $64$ and $16$ feature maps are exploited in first and second layer  respectively. Finally,  $2$ heat maps are used for the  prediction of manipulated and non-manipulated class at the end of decoder network. 
Fig.~\ref{ovFrame} shows the decoder operation of the network. 
At the end of network, we obtain finer representation of spatial maps that indicates the manipulated regions in an image. 

\begin{figure*}[t]
  	\begin{center}
  		\begin{tabular}{c}
  			\includegraphics[height=.18\linewidth]{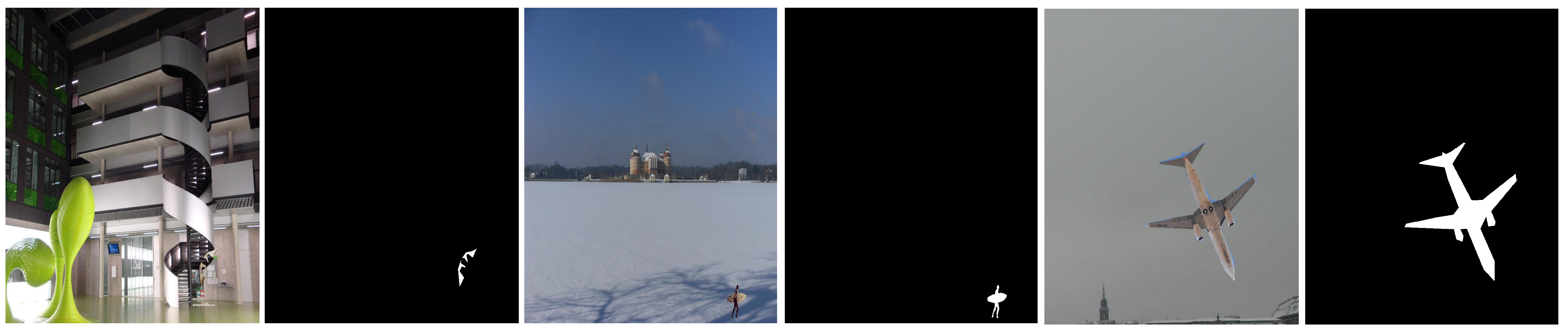} 
  			 \\
  			 (a)\\
  		\includegraphics[height=.18\linewidth]{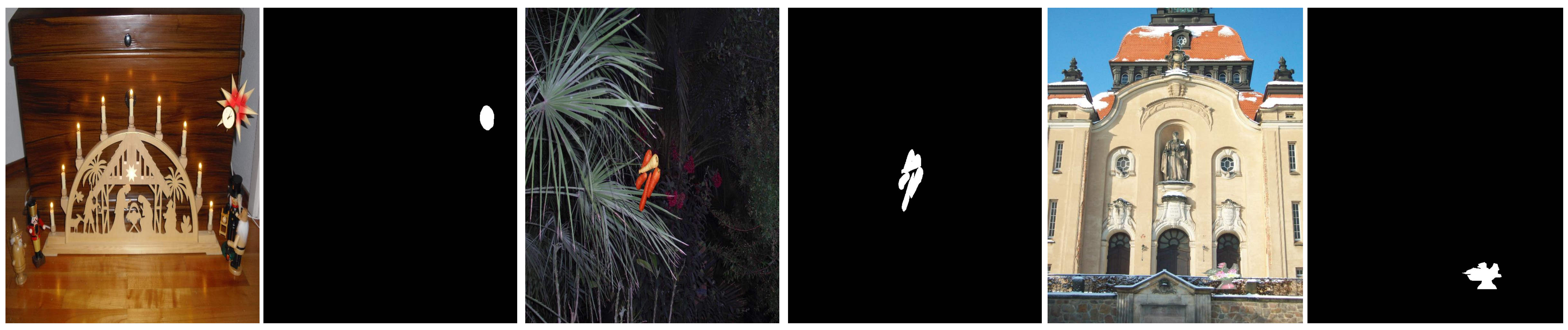} 
  			 \\
  			 (b) \\
  			 \includegraphics[height=.18\linewidth]{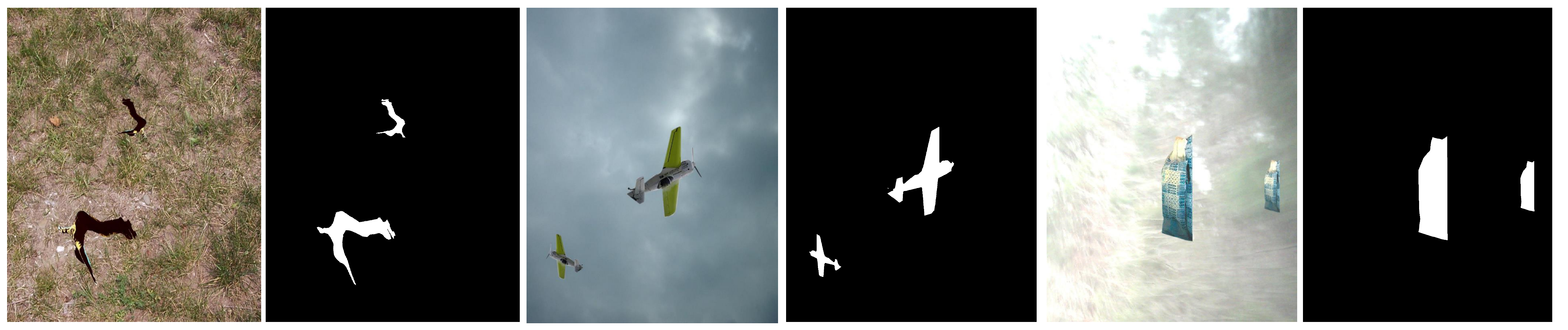}
  			 \\	 (c)\\
  			  \includegraphics[height=.18\linewidth]{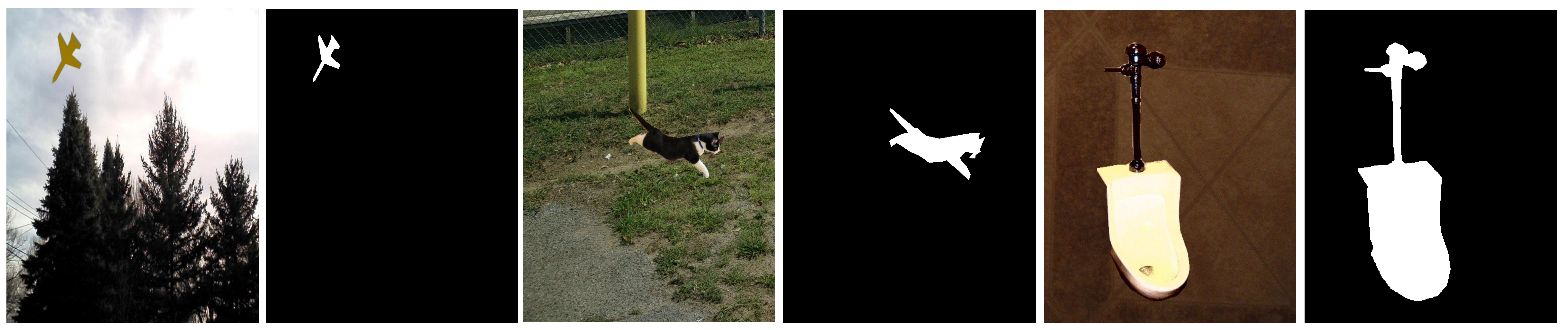}
  			 \\	 (d)\\
  		\end{tabular}
  	\end{center}
  	\vspace{-2mm}
  	\caption{The figures show some manipulated images with corresponding ground-truth masks from synthetic dataset. (a) and (b) show images created from DRESDEN \cite{gloe2010dresden}  dataset. (c) and (d) are the manipulated images created from NIST \cite{2016-nimble-dataset} dataset. }
  	\label{synData}  
  	\vspace{-2mm}
   \end{figure*}

\subsection{Training the Network}

{ \bf Soft-max Layer.}
In order to predict the pixel-wise classification, softmax layer is used at the end of the network. 

Let us denote the  probability distribution over various classes  as $  P(\mathcal{Y}_k) $ which is provided by softmax classifier.
Now, we can predict label by maximizing $ P(\mathcal{Y}_k) $ with respect to $ k $. The predicted label can be obtained by $ \hat{\mathcal{Y}}=\arg\underset{k}\max \ \ P(\mathcal{Y}_k) $.
As we are only interested to predict manipulated pixels against non-manipulated pixels, the value of $k$ would be $2$. Given the predicted mask provided by softmax layer, we can compute the loss that will be used to learn the parameter through back-propagation.

{\bf Training Loss.}
During training, we use cross entropy loss, which is minimized to find the optimal set of parameters of the network.
Let $ \theta $ be the parameter vector corresponding to image tamper localization task. 
So, the cross entropy loss can be computed as
\begin{equation}
\label{loss_s}
\mathcal{L}(\theta)=-\frac{1}{M}\sum_{m=1}^{M}\sum_{n=1}^{N}\mathbbm{1}(\mathcal{Y}^m=n)\log (\mathcal{Y}^m=n|y^m;\theta)
\end{equation}
Here, $M$ and $N$ denote the total number of pixels, and the number of class. $y$ represents the input pixel.
$ \mathbbm{1}(.) $ is an \textit{indicator function}, which equals to $ 1 $ if $ m=n $, otherwise it equals $ 0 $. In our experiment, we observe that weighted cross entropy loss provides better result. It is simply because the imbalance between the number of non-manipulated and manipulated pixels. We put more weight on manipulated pixels over non-manipulated pixels.
In this work, the class weights are inversely proportional to their frequency in the training set. The weights are normalized to lie in between $0$ and $1$.
We use  \textit{adaptive moment estimation (Adam)} \cite{kingma2014adam} optimization technique in order to minimize the loss of the network, shown in Eqn.~\ref{loss_s}.  At each iteration, one mini-batch is processed to update the parameters of the network. 
In order to learn the parameters effectively, we choose the mini-batch very carefully which will be discussed in details in Sec.~\ref{exp}.
After optimizing the loss function over several epochs, we learn  the optimal set of parameters of the network. With these optimal parameters, the network is able to predict pixel-wise classification given a test image.

\begin{table}[h]
\vspace{-2mm}
 \caption{A comparison of common image tampering datasets}
    \label{tab:dataset_comparison}
    \vspace{-2mm}
    \begin{center}
    \begin{tabular}{|c|c|c|}
        \hline
        Data Set & \# image pairs & Avg. Image Size\\
        \hline
        CoMoFod\cite{tralic2013comofod} & $260$ & $512 \times 512$\\
        \hline
        Manip\cite{6301704} & $48$ & $2305 \times 3020$\\
        \hline
        GRIP\cite{7154457} & $100$ & $1024 \times 786$\\
        \hline
        COVERAGE\cite{wen2016} & $100$ & $400 \times 486$\\
        \hline
        \textbf{Synthesized} & $65k$ & $1024 \times 1024$\\
        \hline
    \end{tabular}
    \end{center}
    \vspace{-2mm}
\end{table}
  
\begin{figure*}[t]
  	\begin{center}
  		\begin{tabular}{c}
  			\includegraphics[height=.2\linewidth]{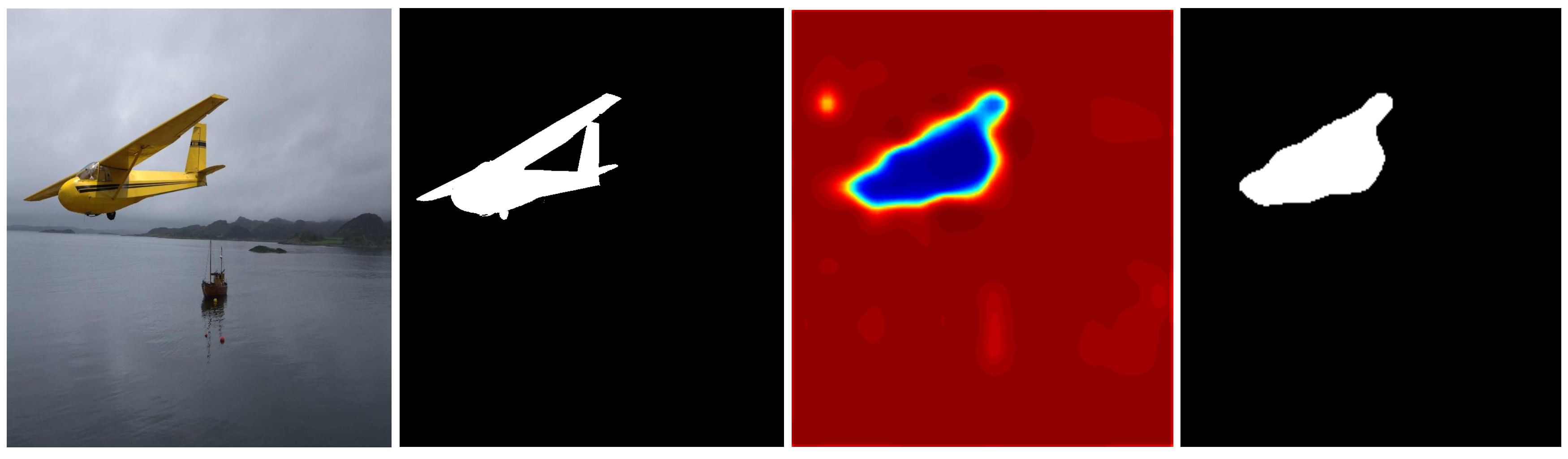} \\
  			(a) \\
  			\includegraphics[height=.2\linewidth]{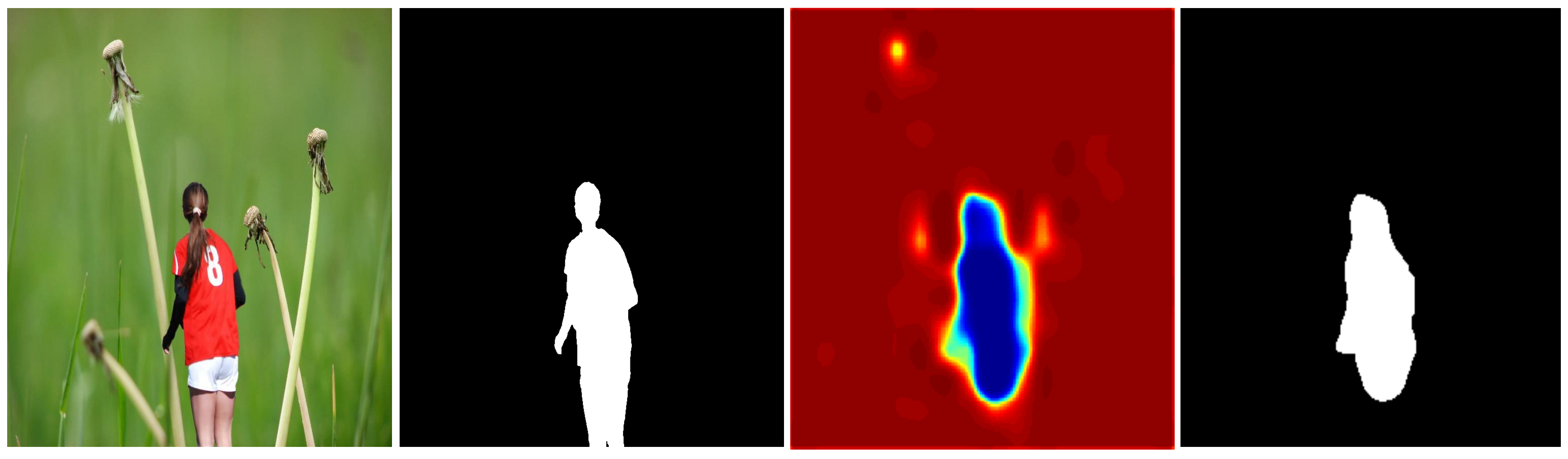} \\
  			 (b)\\
  		\includegraphics[height=.2\linewidth]{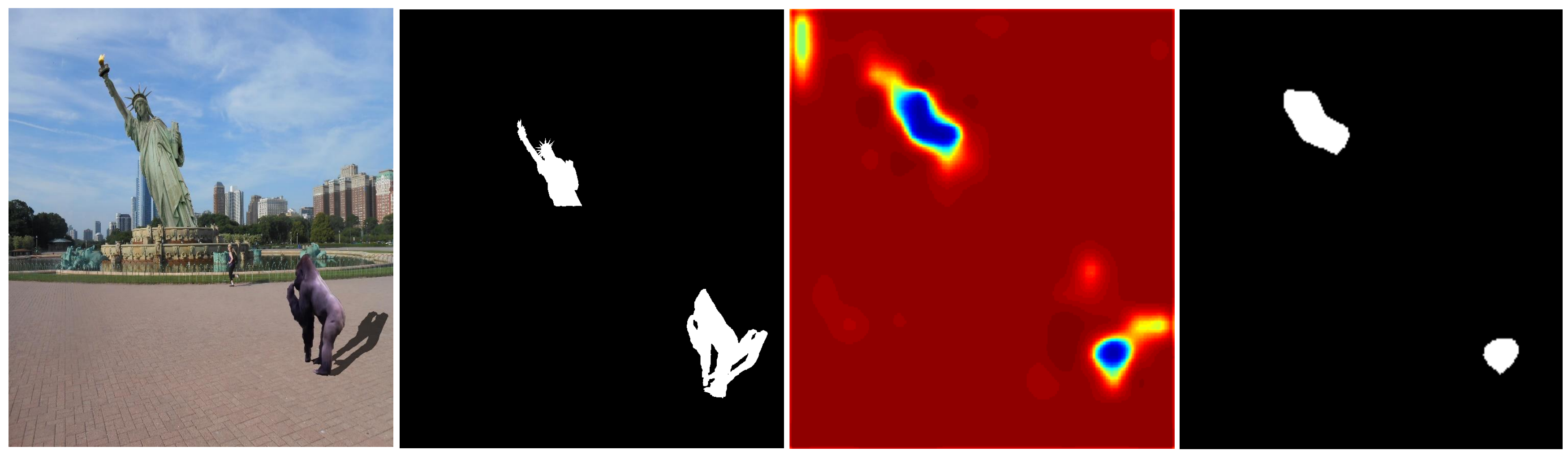} \\
  		(c) \\
  		\includegraphics[height=.2\linewidth]{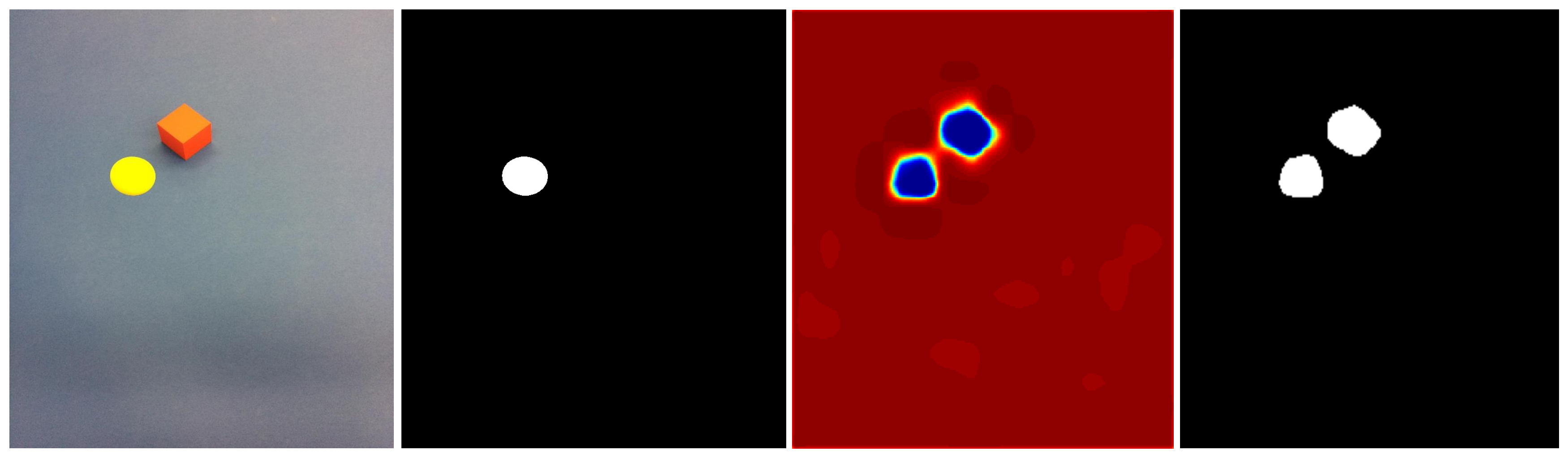} \\
  			 (d)\\
  			 
  		\end{tabular}
  	\end{center}
  	\vspace{-2mm}
  	\caption{This figure illustrates some segmentation results on NIST'16 \cite{2016-nimble-dataset} dataset. First and second columns represent input image and ground-truth mask for tampered region. Third and fourth columns delineate probability heat map and predicted binary mask. 
  	}
  	\label{Seg_res_nc}  
  	\vspace{-2mm}
   \end{figure*}

\begin{figure*}[t]
  	\begin{center}
  		\begin{tabular}{c}
  			\includegraphics[height=.2\linewidth]{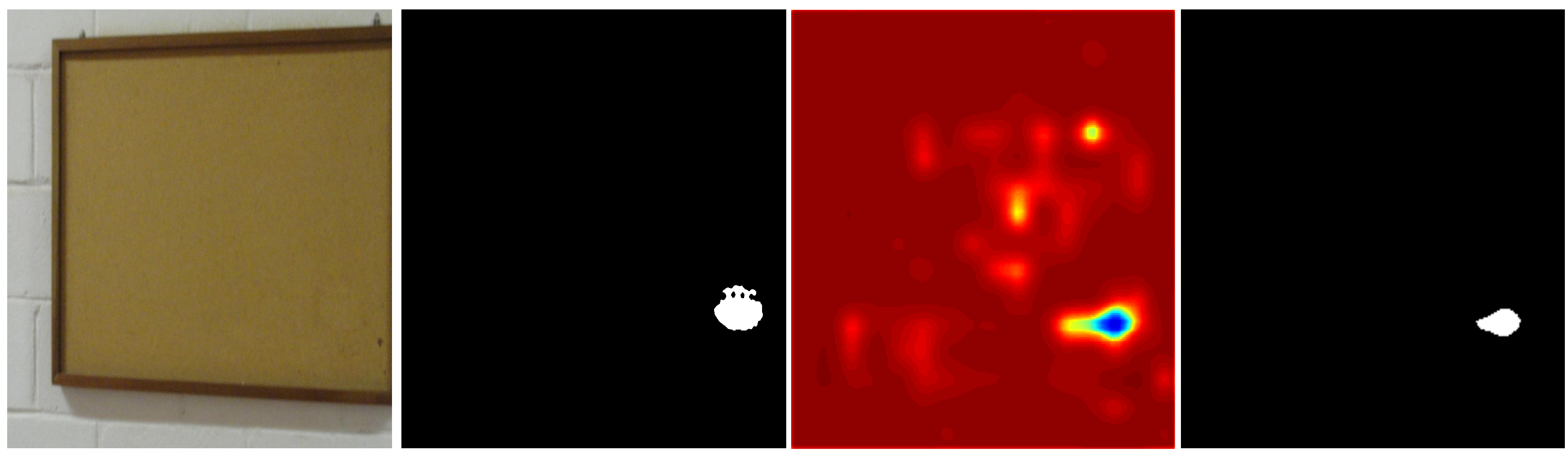} \\
  			(a) \\
  			\includegraphics[height=.2\linewidth]{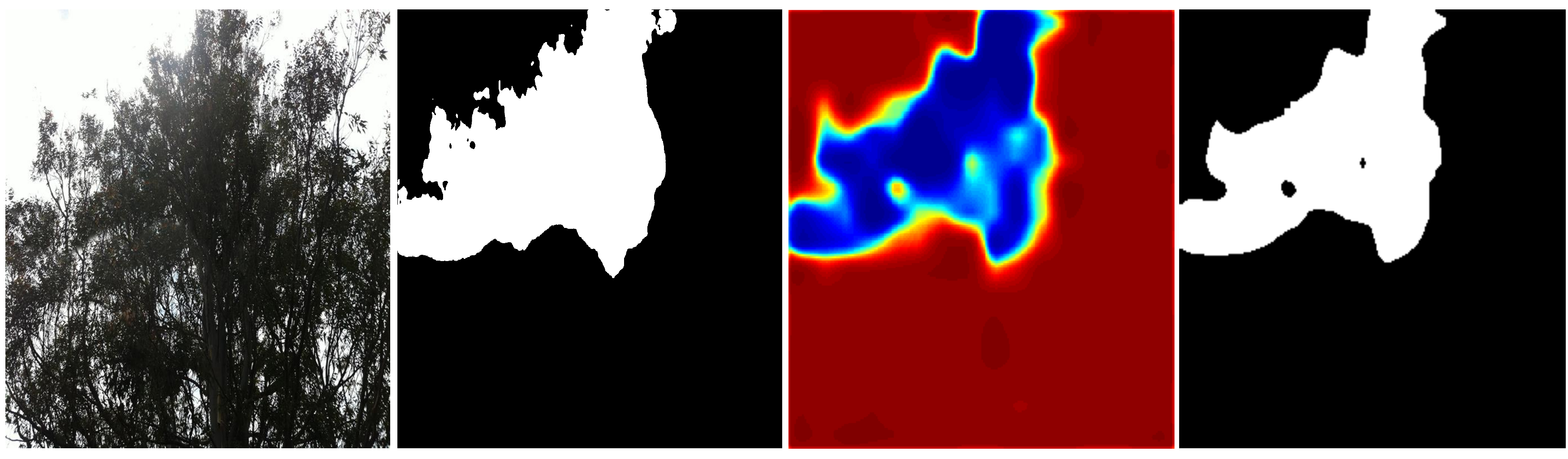} \\
  			 (b)\\
  		\includegraphics[height=.2\linewidth]{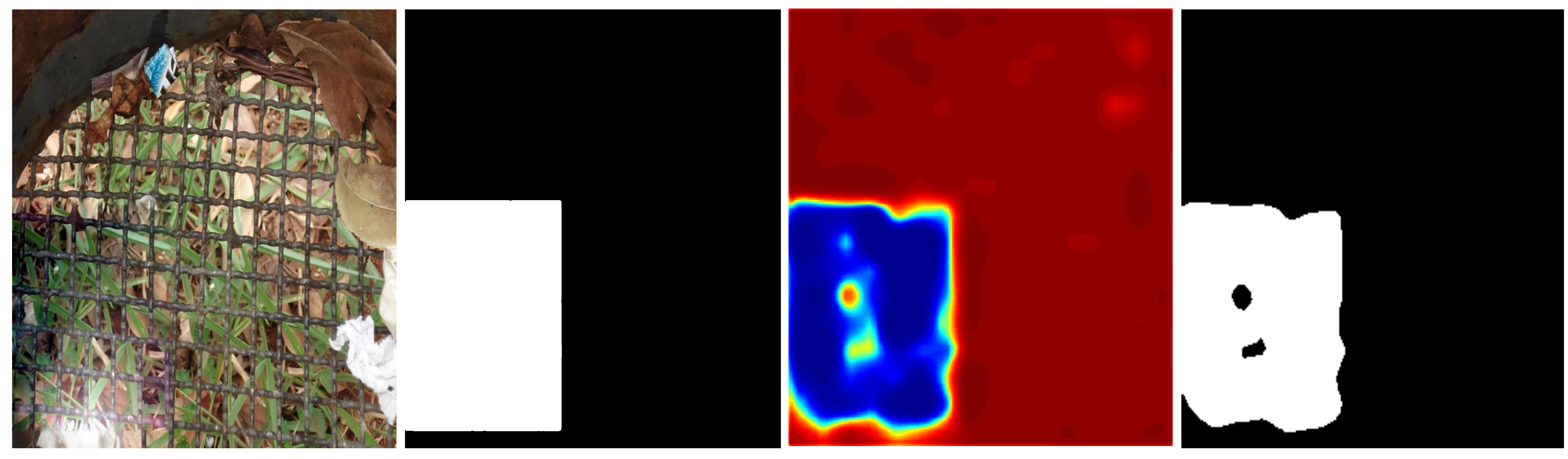} \\
  		(c) \\
  		\includegraphics[height=.2\linewidth]{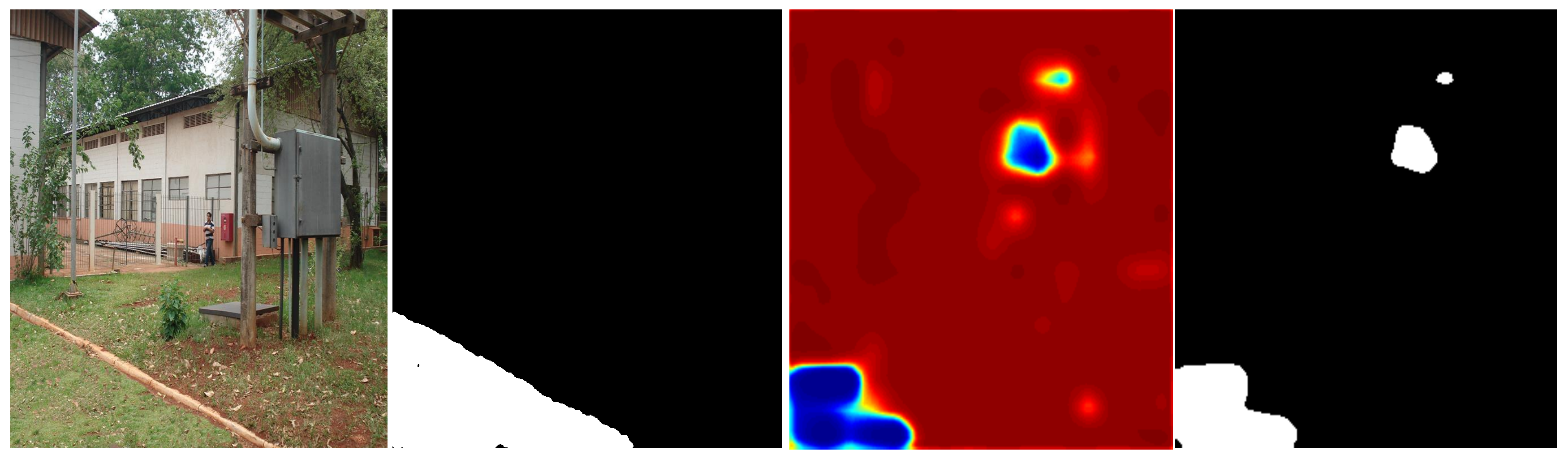} \\
  			 (d)\\
  			 
  		\end{tabular}
  	\end{center}
  	\caption{Some segmentation examples on IEEE Forensics Challenge \cite{2013-ifc-challenge} dataset are shown in this figure. First and second columns are input images and ground-truth masks for manipulated regions. Third and fourth columns demonstrate the probability heatmap and predicted binary mask. 
  	}
  	\label{Seg_res_ieee}  
   \end{figure*}

\begin{figure*}[t]
  	\begin{center}
  		\begin{tabular}{ccc}
  			\includegraphics[height=.2\linewidth]{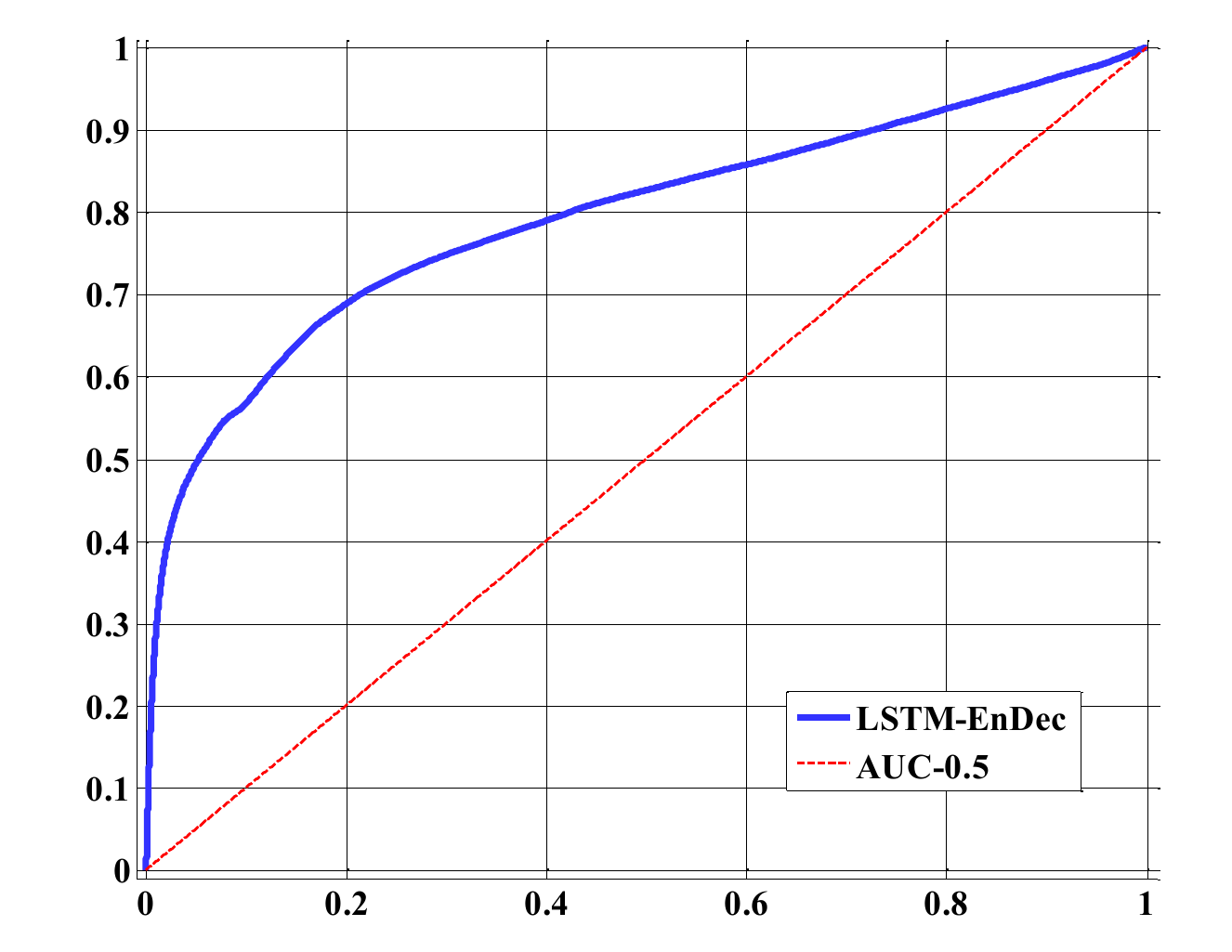} & 
  			\includegraphics[height=.2\linewidth]{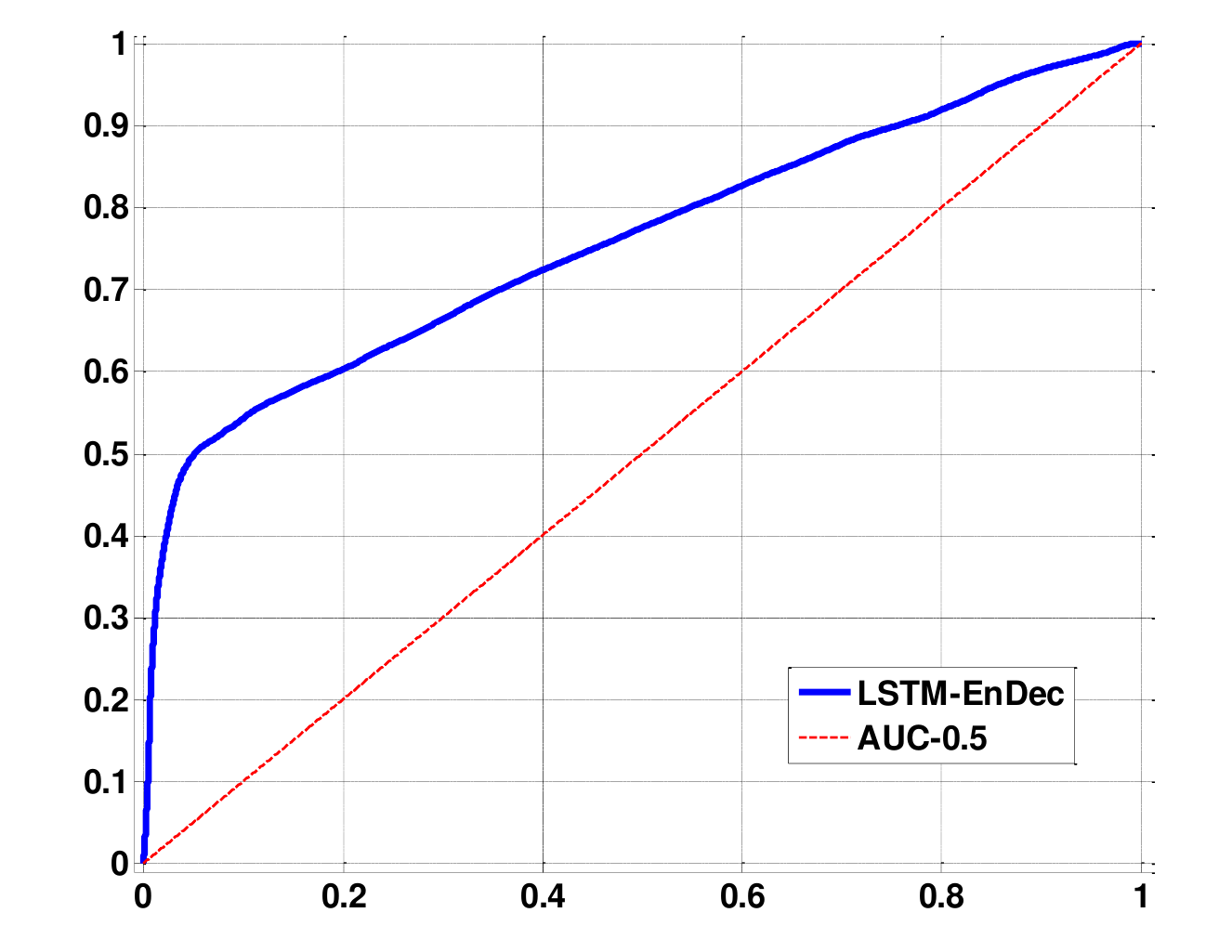}  &
  			\includegraphics[height=.2\linewidth]{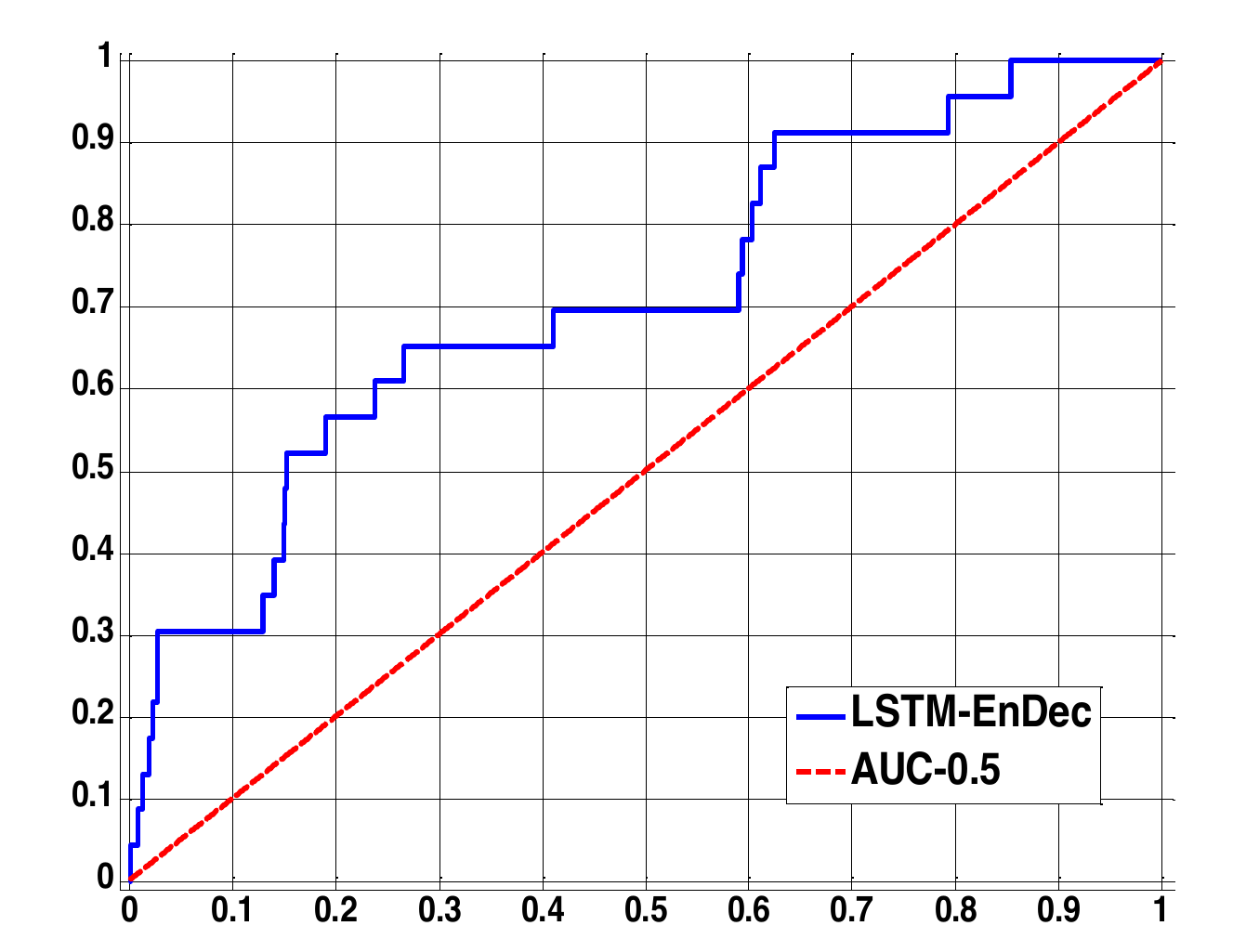}
  			 \\
  			 (a) NIST'16 \cite{2016-nimble-dataset} & (b) IEEE Forensics Challenge \cite{2013-ifc-challenge} & (c) COVERAGE \cite{wen2016}   \\
  	
  		\end{tabular}
  	\end{center}
  	\vspace{-2mm}
  	\caption{The figures demonstrate  ROC plots on NIST'16 \cite{2016-nimble-dataset},  IEEE Forensics Challenge \cite{2013-ifc-challenge} and COVERAGE \cite{wen2016} datasets respectively. Each curve has area under the curve (AUC), which are provided in Table ~\ref{auc_comp_res}.	}
  	\label{roc_curve}  
  	\vspace{-2mm}
  \end{figure*}   
   
\section{Experiments}
\label{exp}
In this section, we demonstrate our experimental results for segmentation of manipulated regions given an image. We evaluate our proposed model on two challenging datasets- NIST'16 \cite{2016-nimble-dataset}, IEEE Forensics Challenge \cite{2013-ifc-challenge} and COVERAGE \cite{wen2016} datasets.

\subsection{Datasets}

\subsubsection{Creation of Synthesized Data}
\label{synthesized}
As deep learning networks are extremely data hungry, there is a need to collect images for training and testing the networks. For training, we will need plentiful examples (usually tens of thousands) of both manipulated and non-manipulated images. Towards this goal, we create approximately $65k$ manipulated images in order to train the proposed network discussed in Sec.~\ref{LSTM-Conv}. This network will be referred to as `Base-Model'.
The `Base-Model' will then be fine-tuned with the NIST'16  \cite{2016-nimble-dataset} and IEEE Forensics Challenge \cite{2013-ifc-challenge} datasets. Below we explain the innovation in the collection of the manipulated image set.

In the synthesized dataset, we have focused on mainly  object splicing (additions/subtractions) manipulation. The major challenge of creating manipulated images was to obtain segmented objects to insert into an image. For this we used the MS-COCO \cite{lin2014microsoft}, which is largely used for object detection and semantic segmentation, to obtain segmented objects across a variety of categories. We extracted the objects from MS-COCO\cite{lin2014microsoft}
images using image masks provided in ground-truth. Finally, these objects are used to create manipulation from the images of DRESDEN \cite{gloe2010dresden}
and NIST'16 \cite{2016-nimble-dataset}. To attempt to emulate a copy-move attack in some cases we spliced multiple version of the same object onto an image, however the difficulty in obtaining segmented object automatically makes it infeasible to perform automated synthesis of copy-move attacks.  Please note that we use only non-manipulated images from NIST'16 dataset to create manipulation. 

To create a new manipulated image, we followed the steps below. \\
(1) For each raw image in the DRESDEN \cite{gloe2010dresden}, we cropped each of the image's corners to extract a $1024 \times 1024$ patch. This method avoids resizing which introduces additional image distortions. \\
(2) For each of these image patches we spliced on six different objects, from the MS-COCO, to create six splice manipulated images. \\
(3) In order to create diverse splicing data, we spliced the same object onto the patch twice with different scaling and rotation factor, while ensuring no overlap as shown in Fig.~\ref{synData}.\\
This entire process was automated allowing us to generate tens of thousands of images in less than a day with no human interaction.
Using the DRESDEN image database as the source of non-manipulated images we were able to produce approximately $40k$ images and an additional $25k$ using the DRESDEN and NIST'16 datasets respectively.
The scale of our data is a hundred fold increase over most datasets that offer similar types of manipulations, which allows us to train a deep learning model. Our synthesized data also has a relatively high resolution. We can see how our dataset compares to similar datasets in table \ref{tab:dataset_comparison}.
With this newly generated data, we trained the `Base-Model'. The base model predicts manipulated region at pixel level given an image.

\begin{figure*}[t]
  	\begin{center}
  		\begin{tabular}{cc}
  			\includegraphics[height=.16\linewidth]{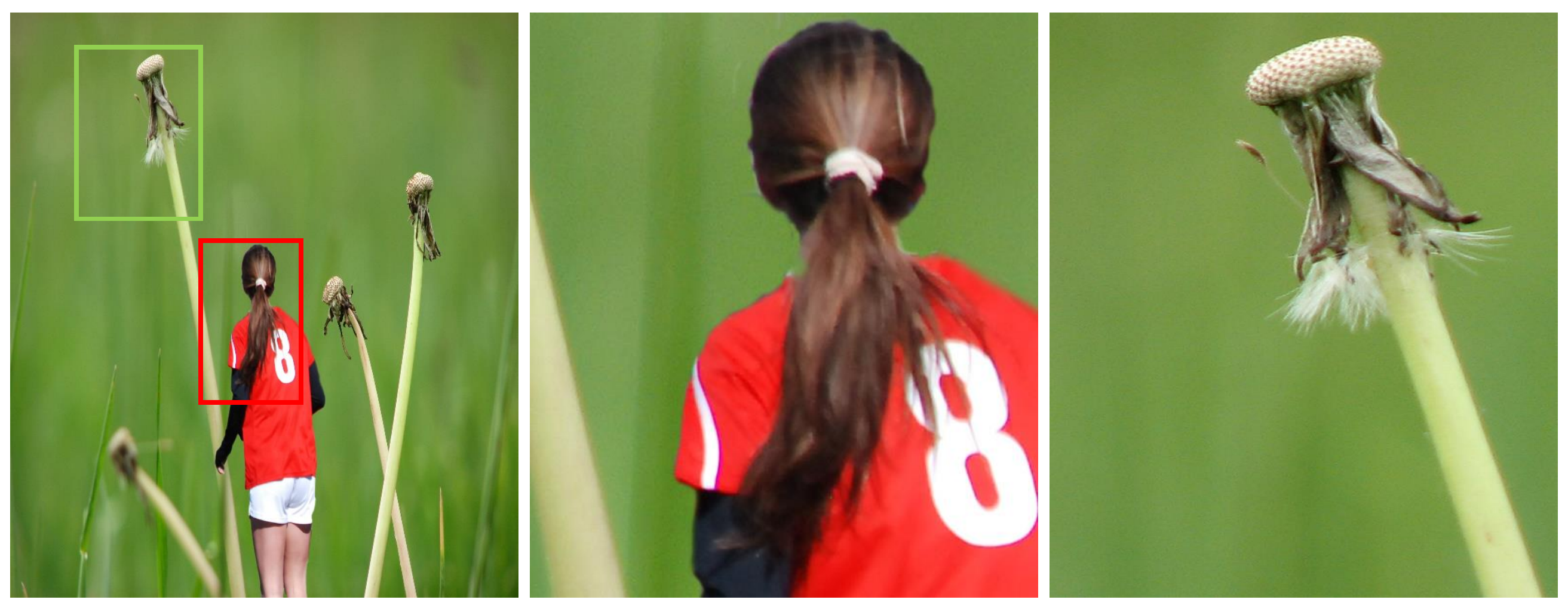} & \includegraphics[height=.16\linewidth]{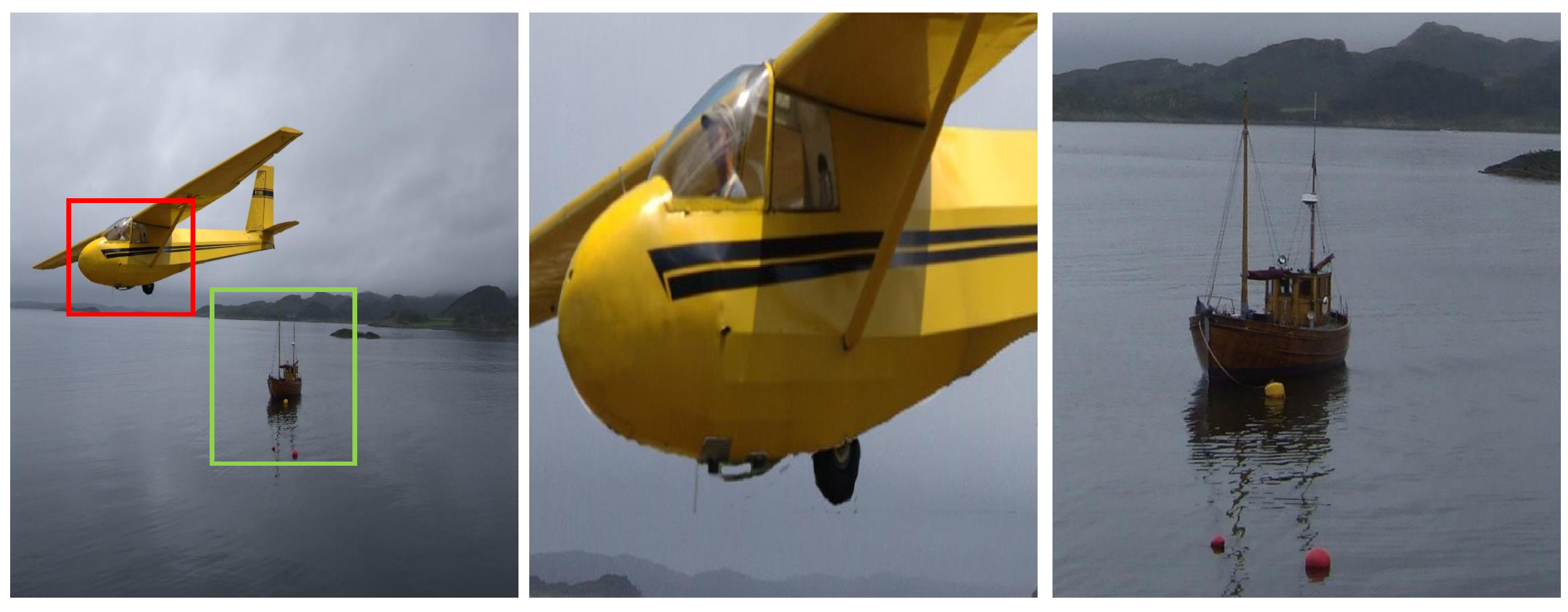}\\
  			(a) & (b) \\
  			\includegraphics[height=.16\linewidth]{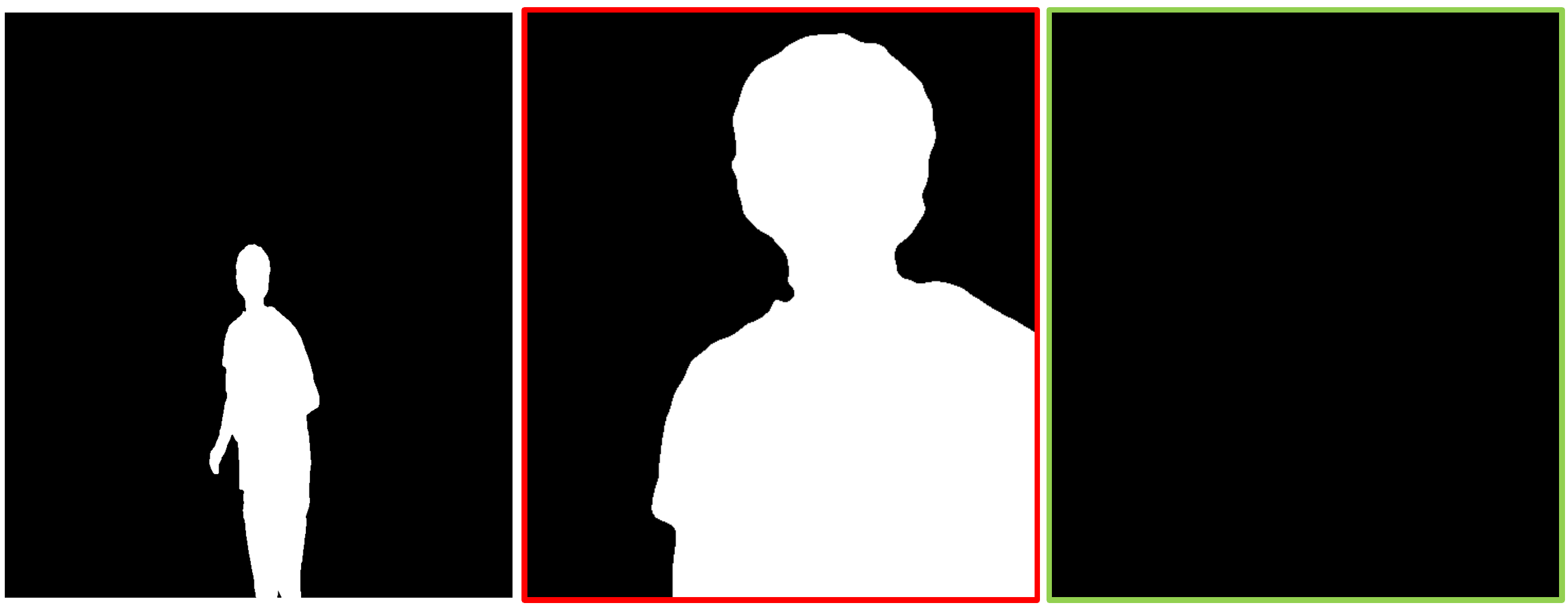} & \includegraphics[height=.16\linewidth]{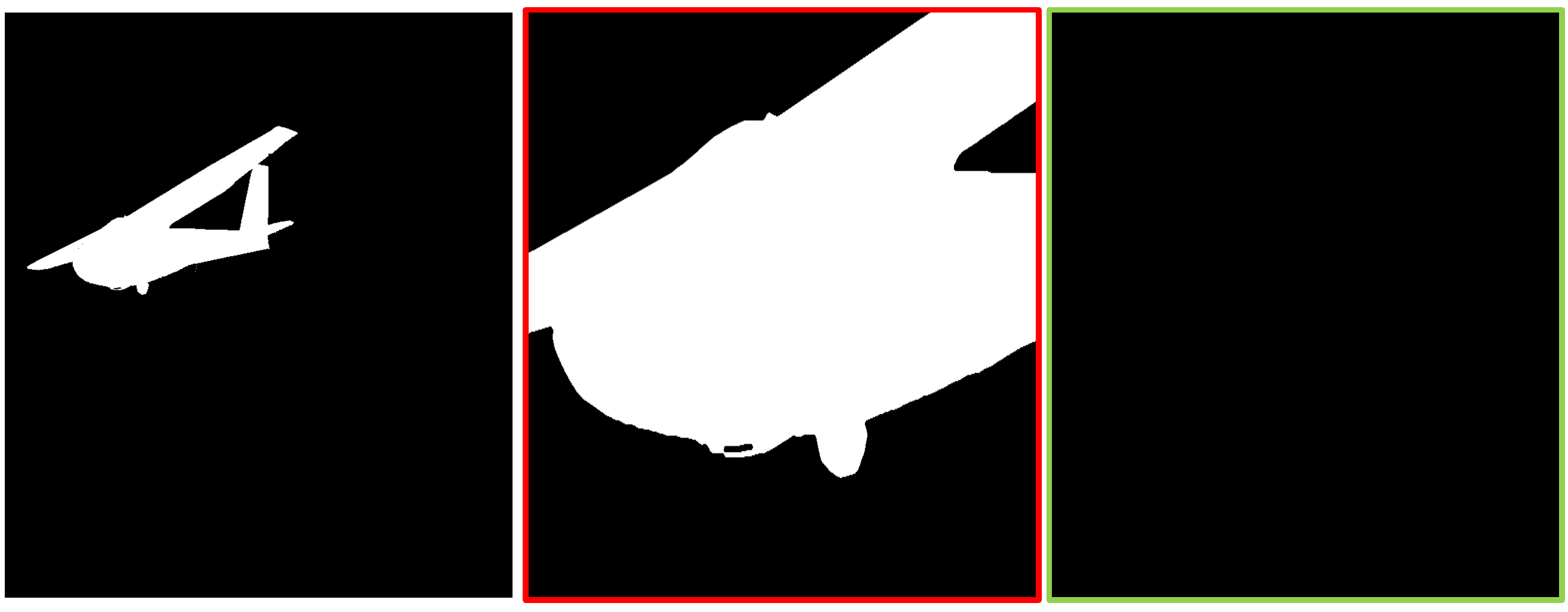}\\
  			 (c) & (d)\\
  			\includegraphics[height=.16\linewidth]{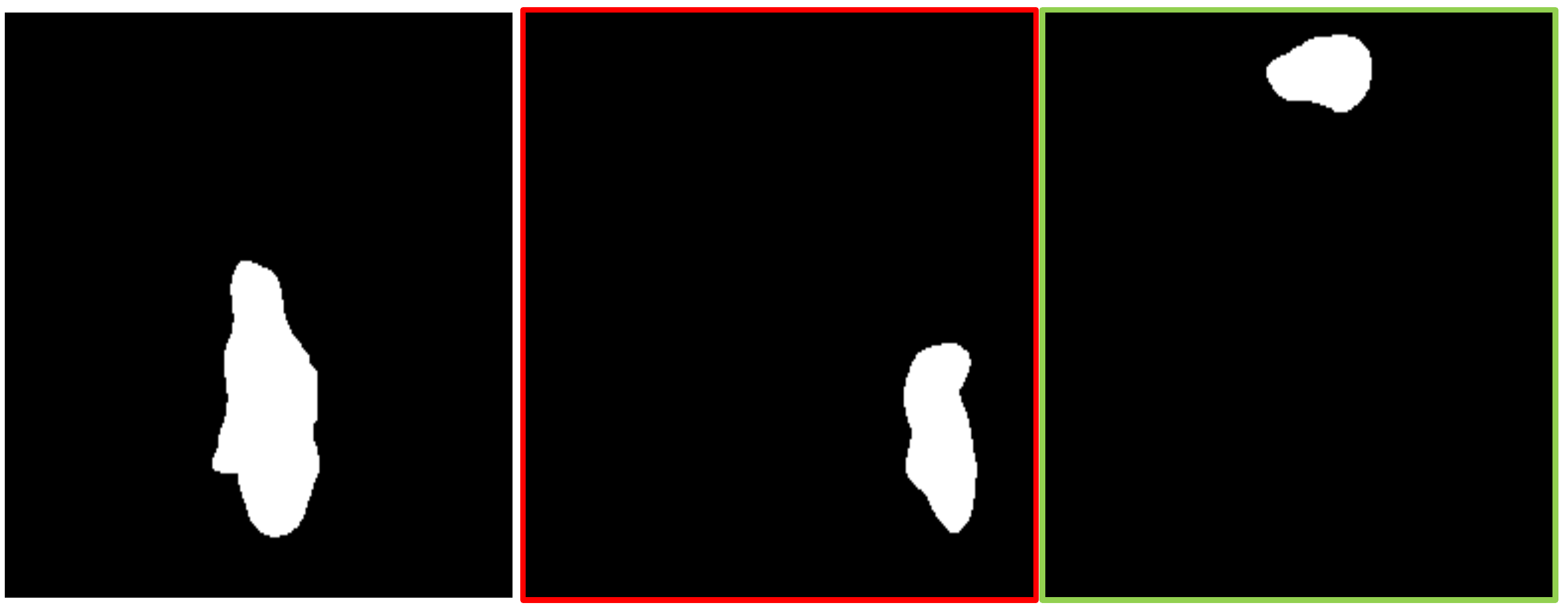} & \includegraphics[height=.16\linewidth]{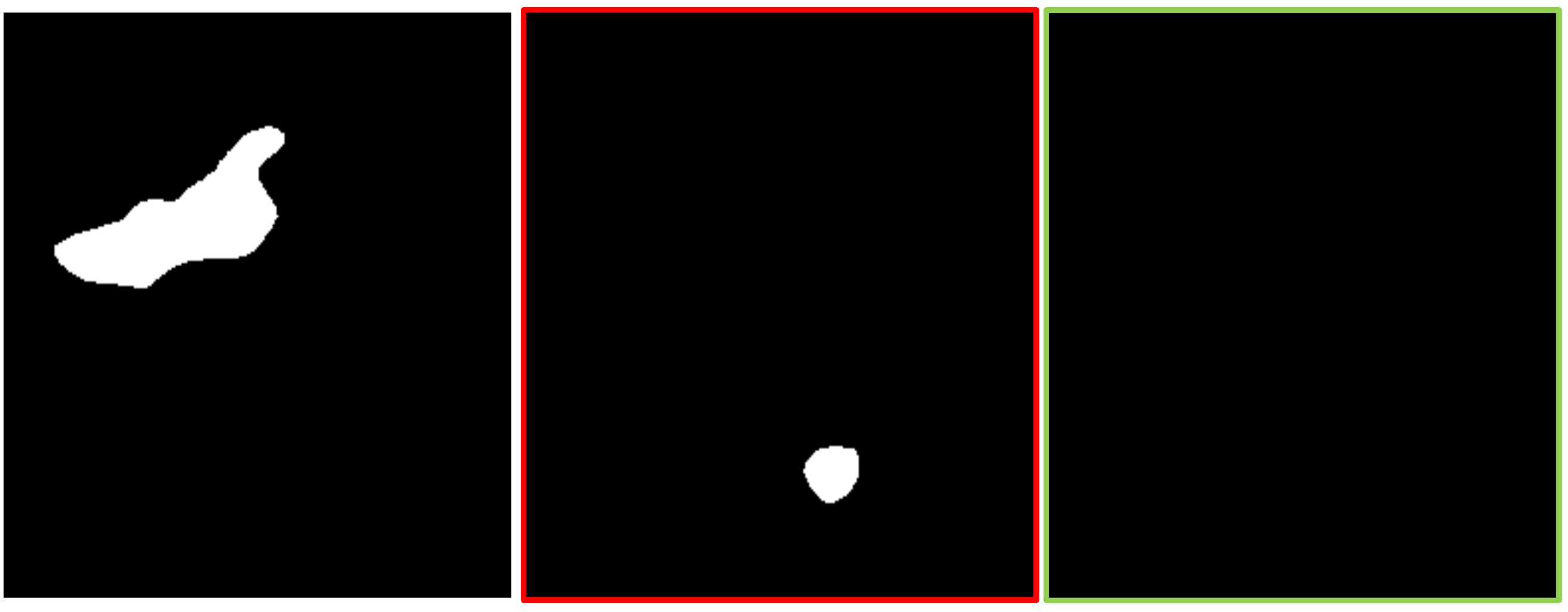}\\
  		(e) &  (f)\\
  			 
  		\end{tabular}
  	\end{center}
  	\vspace{-2mm}
  	\caption{This figure demonstrates the segmentation performance with patches as input on NIST'16 \cite{2016-nimble-dataset} dataset. First column of (a)  represents the input image. Second and third columns of (a) delineate the patches as shown in the bounding boxes of input image (first column). Figures (c,d) and (e,f) are corresponding ground-truth mask and 
      predicted binary mask. 
  	}
  	\label{qual_pat_res}  
  	\vspace{-2mm}
   \end{figure*}
   
\subsubsection{Dataset Preparation}
\label{dataprep}
In order to evaluate our model, we chose three datasets which provided ground-truth mask for manipulated regions. 
NIST'16 \cite{2016-nimble-dataset} is a very challenging dataset, which includes three main types of manipulation - (a) copy-clone, (b) removal, and (c) splicing.
This recently released dataset includes images, which are tampered in a sophisticated way to beat current state-of-the-art detection techniques. 
We also show our results on the IEEE Forensics Challenge  \cite{2013-ifc-challenge} dataset which provides ground-truth mask for manipulation. As manipulated regions are small in number compared to non-manipulated regions, we also perform data augmentation in order to get rid of bias in training. 
In addition, we choose COVERAGE \cite{wen2016} to demonstrate the performance of our proposed model on copy-move manipulation.

In data preparation, we first split the whole image dataset into three subsets- training ($ 70\% $), validation ($ 5\% $) and testing ($ 25\% $). These subsets are chosen randomly.  
In order to increase the training data, we extract bigger patches from the four corners of the image. One additional patch is also extracted from center location of the image.
We crop  patches with size $1024 \times 1024$ from NIST'16 \cite{2016-nimble-dataset} training images to optimize the parameters of our architecture.  The spatial resolution of IEEE Forensics Challenge  \cite{2013-ifc-challenge} dataset is comparatively low. So, we extract $512 \times 512$ size of patches for IEEE Forensics Challenge  \cite{2013-ifc-challenge} dataset.   These newly generated images usually contain partial manipulated objects when compared to original images. We only perform data augmentation on training set, not in validation and test set.
As the image and corresponding ground-truth mask are the same size, we can easily generate the ground-truth masks for the extracted image patches. With these newly generated ground-truth masks and patches, we train the whole network end-to-end. 
For COVERAGE \cite{wen2016} dataset, we do not train the proposed network due to small number of samples.

\subsection{Experimental Analysis}
In this section, we will discuss the implementation and evaluation criterion of our model. We also compare our model with different state-of-the-art methods for segmentation of manipulated regions.

\textbf{Implementation Details.}
We implement our proposed framework in TensorFlow. In order to expedite our computational load, we utilize multi-GPU setting. We use two NVIDIA Tesla $K80$ GPUs  to perform different sets of  experiments, which will be discussed next.

\textbf{Evaluation Criterion.}
In order to evaluate our model, we use pixel-wise accuracy and  receiver operating characteristic (ROC) curve. ROC curve measures the performance of binary classification task by varying the threshold on prediction score.
The area under the ROC curve (AUC) is computed from the ROC curve that measures
the distinguishable ability of a system for binary classification. The AUC value typically lies in between $0$ and $1.0$. The AUC with $1.0$ is sometimes referred as perfect system (no false alarm).

\textbf{Experimental Setup.}
In this paper, we setup few experiments to evaluate our proposed architecture. They are (1)  performance of the proposed model,  (2) performance with different baseline methods, (3) comparison against existing state-of-the-art approaches, (4) ROC curve,  (5) qualitative analysis, and (6) impact of global context.

\begin{table}[t]
\begin{center}
\vspace{-2mm}
\caption{The table shows the pixel-wise accuracy on  NIST'16 \cite{2016-nimble-dataset}, IEEE Forensics Challenge \cite{2013-ifc-challenge} and COVERAGE \cite{wen2016} datasets for image tamper segmentation. }
\label{seg_res}
\vspace{-3mm}
\begin{tabular}{|l|c|c|c|}
\hline
Methods & NIST'16 \cite{2016-nimble-dataset} & IEEE \cite{2013-ifc-challenge}& COVERAGE \cite{wen2016}\\
\hline
FCN \cite{long2015fully} & $74.28\%$ &-- & --\\
\hline
Encoder-Decoder \cite{badrinarayanan2017segnet}  & $82.96\%$ & -- & -- \\
\hline
J-Conv-LSTM-Conv \cite{bappy2017exploiting} &  $84.60\%$  & $77.67\%$ & $81.14\%$\\
\hline
LSTM-EnDec-Base & $91.36\%$  & $88.24\%$ & $88.76\%$\\ 
\hline
LSTM-EnDec &  ${\bf 94.80}\%$  & $ {\bf 91.19}\%$& -- \\ 
\hline
\end{tabular}
\end{center}
\end{table}
{\bf Baseline Methods:}   In this section, we will introduce some baseline methods. We implement and compare against these methods. The various baseline methods are described below.
\newline 
$ \diamond $ \textit{FCN} : Fully convolutional network as presented in \cite{long2015fully}.\\
$ \diamond $ \textit{J-Conv-LSTM-Conv}: This method utilizes LSTM network and convolutional layers for segmentation as in \cite{bappy2017exploiting}. \\ 
$ \diamond $ \textit{Encoder-Decoder}: This method utilizes convolutional network as encoder and deconvolution as decoder, proposed in \cite{badrinarayanan2017segnet}. \\
$ \diamond $ \textit{EnDec}: Similar to encoder-decoder \cite{badrinarayanan2017segnet} with upsampling factor of $4$ in deconvolution.\\
$ \diamond $ \textit{LSTM-EnDec-Base}: Proposed architecture as shown in Fig.~\ref{ovFrame} trained on Synthesized dataset discussed in Sec.~\ref{synData}\\
$ \diamond $ \textit{LSTM-EnDec}:Finetuned model of proposed architecture  as shown in Fig.~\ref{ovFrame}
\\

\subsubsection{Performance of the Proposed Model.} 
We test our proposed model on three datasets- NIST'16 \cite{2016-nimble-dataset}, IEEE Forensics Challenge \cite{2013-ifc-challenge} and COVERAGE \cite{wen2016}. We first train our model with synthesized data (discussed in Sec.~\ref{synData}). 
We refer this model as `LSTM-EnDec-Base' model.  The LSTM-EnDec-Base model is finetuned with training sets from  NIST'16 \cite{2016-nimble-dataset}, IEEE Forensics Challenge \cite{2013-ifc-challenge} datasets. We obtain two finetuned models for NIST'16 and IEEE Forensics Challenge datasets respectively. 
As the number of images in COVERAGE \cite{wen2016} dataset is small, we do not perform any finetuning.
Table~\ref{seg_res} shows pixel-wise classification accuracy on  segmentation task. 
`LSTM-EnDec-Base' model learns good discriminative properties between manipulated vs non-manipulated pixels. Finally, finetuning this `LSTM-EnDec-Base' model provides a boost in performance for labeling tamper class at pixel level.
From the table, we can see that proposed model `LSTM-EnDec'  outperforms  `LSTM-EnDec-Base' model by $3.44\%$, and $2.95\%$ on NIST'16 \cite{2016-nimble-dataset}, IEEE Forensics Challenge \cite{2013-ifc-challenge} datasets respectively.

\subsubsection{Performance with Different Baseline Methods.} 
In semantic segmentation, some recent architectures such as fully convolutional netowork (FCN) \cite{long2015fully} and Encoder-Decoder (SegNet) \cite{badrinarayanan2017segnet} have successfully exploited. In this paper, we  implement and train these deep architectures with image manipulation data to compare the performance of our model.
We can see from Table.~\ref{seg_res} that convolutional neural network based model such as FCN, and SegNet does not perform well compared to proposed architecture for tamper localization.
It is  because these models try to learn the visual concept/feature from an image whereas manipulation of an image does not leave any visual clue. We empirically observe that FCN and SegNet prone to misclassify for copy-clone and object removal type of manipulations. 
LSTM-EnDec surpasses FCN and Encoder-Decoder network by $20.52\%$ and $11.84\%$  on NIST'16 \cite{2016-nimble-dataset} as shown in Table.~\ref{seg_res}. 
We also compare against the segmentation framework for tamper localization (J-Conv-LSTM-Conv) presented in \cite{bappy2017exploiting}. The proposed network outperforms J-Conv-LSTM-Conv by large margin. The advantage of our proposed model over J-Conv-LSTM-Conv is that proposed model can learn larger context by exploiting correlation between patches. On the other hand, J-Conv-LSTM-Conv is limited to correlate between different blocks of a patch. 
The exploitation of both LSTM network with resampling features and spatial features using encoder, helps the overall architecture to learn manipulations better.

\begin{table}[t]
\vspace{-2mm}
\caption{AUC Comparison against existing approaches on NIST'16 \cite{2016-nimble-dataset}, IEEE \cite{2013-ifc-challenge} and COVERAGE \cite{wen2016} datasets. }
\label{auc_comp_res} 
\vspace{-3mm}
\begin{center}
\begin{tabular}{|l|c|c|c|}
\hline
Methods & NIST'16 \cite{2016-nimble-dataset}& IEEE \cite{2013-ifc-challenge} & COV. \cite{wen2016} \\
\hline
DCT Histograms~\cite{lin2009fast} & $0.545$ & -- & --\\
\hline
ADJPEG~\cite{bianchi2012image}  & $ 0.5891$ & -- & --\\
\hline
NADJPEG~\cite{bianchi2012image} & $0.6567$  & -- & --\\ 
\hline
PatchMatch~\cite{cozzolino2015efficient} & $0.6513$ & -- & -- \\ 
\hline
Error level analysis~\cite{luo2010jpeg} & $ 0.4288$ & --  & --\\ 
\hline
Block Features~\cite{li2009passive} &  $0.4785$ & -- & --\\ 
\hline
Noise Inconsistencies~\cite{mahdian2009using} &  $0.4874$ & -- & --\\
\hline 
J-Conv-LSTM-Conv \cite{bappy2017exploiting} &  $0.7641$ & $0.7238$ & $0.6137$\\ 
\hline
LSTM-EnDec &  ${\bf 0.7936}$ & ${\bf 0.7577}$ &  ${\bf 0.7124}$\\
\hline
\end{tabular}
\end{center}
\end{table}

\subsubsection{Comparison against Existing Approaches.}
Some of the tamper localization techniques include DCT Histograms~\cite{lin2009fast}, ADJPEG~\cite{bianchi2012image}, NADJPEG~\cite{bianchi2012image}, PatchMatch~\cite{cozzolino2015efficient}, Error level analysis~\cite{luo2010jpeg}, Block Features~\cite{li2009passive}, and Noise Inconsistencies~\cite{mahdian2009using}. Table.~\ref{auc_comp_res}.
shows the performance of these state-of-the-art methods for image tamper localization. 
From the table, we can observe that our framework outperforms other existing methods by large margin on NIST'16 \cite{2016-nimble-dataset} dataset.
In our proposed network, resampling features are exploited to predict manipulated regions. To understand the effect of resampling features in the proposed architecture, we run an experiment without LSTM network  and resampling features, which is represented as Encoder-Decoder network in Table.~\ref{seg_res}. As can be seen in Table.~\ref{seg_res}, the proposed model LSTM-EnDec outperforms Encoder-Decoder by large margin ($11.84\%$)  on NIST'16 \cite{2016-nimble-dataset} dataset. 

 We also compare against \cite{zhou2018learning} where a two-stream Faster R-CNN network has been exploited to detect manipulated regions.
 \cite{zhou2018learning} utilized bounding box to coarsely localize manipulated objects. In contrast, we segment out manipulated regions by classifying  a pixel (manipulated/non-manipulated). Since our model does not provide bounding boxes, we exploit contour approximation method to predict a bounding box on the segmentation maps produced by the proposed model.
 We evaluate the performance of our method 
 in terms of average precision (AP) on NIST'16 \cite{2016-nimble-dataset} dataset.   
 We also generate ground-truth bounding boxes on ground-truth binary masks using contour approximation method. In some cases, the proposed method falsely classifies manipulated pixels, and the contour approximation method puts a bounding box around these small false positive pixels. In order to reduce false positive bounding boxes, we use a threshold on the area of rectangle box. In our  case, we eliminate the bounding box which has area under $64$. As a result, we observe significant improvement in AP score. The AP score rises from $0.825$ to $0.923$.
 The AP score of \cite{zhou2018learning} is $0.934$.
 From the above discussion, we can see that the proposed model achieves comparable results to \cite{zhou2018learning} even though the network do not predict a bounding box as output.
 
 \subsubsection{ROC Curve.}
Figs.~\ref{roc_curve}(a,b) show the ROC plots for image tamper localization, on NIST'16 \cite{2016-nimble-dataset},  IEEE Forensics Challenge \cite{2013-ifc-challenge}, and COVERAGE \cite{wen2016} datasets respectively. These ROC curves measure the performance of binary pixel classification whether a pixel is manipulated or not. 
We also provide the area under the curve (AUC) results in Table~\ref{auc_comp_res}.
Our model achieves AUC of $\textbf{0.7936}$, $\textbf{0.7577}$ and $\textbf{0.7124}$ on NIST'16, IEEE Forensics, and COVERAGE datasets respectively. From the ROC curves as shown in Figs.~\ref{roc_curve}(a), ~\ref{roc_curve}(b) and ~\ref{roc_curve}(c), we can see that the proposed network  classifies tampered pixels with high confidence.

\subsubsection{Qualitative Analysis of Segmentation.}
In Figs.~\ref{Seg_res_nc} and ~\ref{Seg_res_ieee}, we provide some examples showing  segmentation results produced by the proposed network. Fig.~\ref{Seg_res_nc} shows segmentation results on NIST'16 \cite{2016-nimble-dataset} dataset. Segmentation results for IEEE Forensics Challenge \cite{2013-ifc-challenge} dataset are illustrated in Fig.~\ref{Seg_res_ieee}. We also provide probability heat map for localizing tampered region as shown in third column of Figs.~\ref{Seg_res_nc} and ~\ref{Seg_res_ieee}. As we can see from the Figs.~\ref{Seg_res_nc} and ~\ref{Seg_res_ieee}, the predicted mask can locate manipulated regions from an image with high probability. The boundary of tampered objects is affected in the segmentation results as shown in  Fig.~\ref{Seg_res_nc} (third column), the underlying reason being that image boundaries are smooth (blurred) for NIST'16 \cite{2016-nimble-dataset}  dataset. However, our proposed network can still localize precisely with higher overlap compared to ground-truth mask.

\subsubsection{Impact of Global Context.} 
In our framework, we consider images as input so that the network can exploit global context. In order to observe the effectiveness of global context, we run an experiment where we consider patches as input to the network instead of images. Fig.~\ref{qual_pat_res} illustrates the segmentation results with respond to the input patches. From the figure, we can see that the network can localize more precisely given an image. On the other hand, the precision of localization degrades for smaller patch as the patch misses the broader context. In case of manipulated patch as shown in Figs.~\ref{qual_pat_res}(a) and ~\ref{qual_pat_res}(b) (middle column), proposed network detects the part of the manipulated objects. For example, \textit{digit of the person's dress} and \textit{wheel of a plane} are identified as manipulated as shown in  Figs.~\ref{qual_pat_res}(e) and ~\ref{qual_pat_res}(f) respectively. For the patch with non-manipulated pixels, the network may provide false alarm sometimes as demonstrated in Figs.~\ref{qual_pat_res}(e) (third column). From this study, we can conclude that global context helps analyzing the manipulated images.

\section{Conclusion}

In this paper, we present a deep learning based approach to semantically segment manipulated regions in a tampered image. 
In particular, we employ a hybrid CNN-LSTM model that effectively classifies manipulated and non-manipulated regions. 
We exploit CNN architecture to design an encoder network that provides spatial feature maps of manipulated objects. Resampling features of the patches are incorporated in LSTM network to observe the transition  between manipulated and non-manipulated patches. 
Finally, a decoder network is used to learn the mapping from encoded feature maps to binary mask. Furthermore, we also present a new synthesized dataset which includes large number of images. This dataset could be beneficial to media forensics community, especially if one wants to train a deep network.
Our detailed experiments showed that our approach could efficiently segment various types of manipulations including copy-move, object removal and splicing.

\vspace{-1mm}\section{Acknowledgement} \vspace{-1mm} \small This research was developed with funding from the Defense Advanced Research Projects Agency (DARPA). The views, opinions and/or findings expressed are those of the author and should not be interpreted as representing the official views or policies of the Department of Defense or the U.S. Government. 
The paper is approved for public release, distribution unlimited. 
{\small
	\bibliographystyle{ieee}
	\bibliography{ucr,forensics-mc,egbib}
}

\IEEEdisplaynontitleabstractindextext
\IEEEpeerreviewmaketitle

\end{document}